\documentclass[letterpaper]{article}
\usepackage[preprint]{aaai2027}
\usepackage[hyphens]{url}
\usepackage{graphicx}
\usepackage{natbib}
\usepackage{caption}
\usepackage{amsmath,amssymb,amsthm}
\usepackage{booktabs}
\usepackage{multirow}
\usepackage{array}

\theoremstyle{definition}
\newtheorem{definition}{Definition}
\newtheorem{example}{Example}

\title{Pairwise Logical Selection of Enthymeme Completions under
Semantic-Link Uncertainty}
\author{
Xuyao Feng\textsuperscript{\rm 1,*},
Antonis Bikakis\textsuperscript{\rm 2}
}
\affiliations{
\textsuperscript{\rm 1}Department of Computer Science,
University College London\\
\textsuperscript{\rm 2}Department of Information Studies,
University College London\\
London, United Kingdom\\
\textsuperscript{\rm *}Corresponding author: Xuyao Feng\\
xuyao.feng.20@ucl.ac.uk, a.bikakis@ucl.ac.uk
}

\begin{document}
\maketitle
\begin{abstract}
Arguments often omit premises or claims, forming enthymemes. We study pairwise
logical selection between two candidates for the omitted component. Natural
language processing methods can identify or generate candidates but often do
not expose how the selected candidate completes the inference. Logic-based
approaches make this inference explicit but usually assume that the required
formulae and background knowledge are available. A prior neuro-symbolic
pipeline translated the stated text and two generated missing-premise
candidates into propositional formulae and tested each candidate independently
with a SAT solver; candidates with the same entailment status remained tied
even when the task required a single selection.
We extend the pipeline to missing-claim selection and replace binary entailment
outcomes with logical-resistance scores for pairwise comparison. We define
\textsc{Top-Link}, which uses weighted Partial MaxSAT to compute logical
resistance under a single configuration of highest-confidence semantic links.
We then introduce \emph{Possible-World Atom-Link Formalization} (PWAL), which
keeps translated formulae fixed and marginalizes logical resistance over
alternative cross-formula semantic-link configurations.
We evaluate five tasks: ARCT and a CDED-derived task for missing-premise
selection; iDebate- and AAE2-derived tasks for missing-claim selection; and
\(\alpha\)NLI for abductive hypothesis selection. Relative to
\textsc{Top-Link}, PWAL raises strict accuracy by
\(2.95\)--\(30.86\) percentage points and reduces tie rates by
\(4.57\)--\(58.00\) percentage points on all five tasks. When ties receive half
credit, accuracy still increases by \(0.45\)--\(6.04\) percentage points. For
each comparison, PWAL records the translated formulae, sampled link
configurations, and resistance components, providing a transparent trace of
how the scores are computed.
\end{abstract}

\section{Introduction}

An enthymeme leaves a premise or claim unstated. For example, the premise
\emph{the weather report predicts rain} does not entail the claim \emph{you
should take an umbrella} without an implicit premise connecting rain to the
need for an umbrella. Recovering the omitted component makes the inference
explicit and permits logical evaluation. We study pairwise logical selection
between two supplied candidate completions, where the omitted component is
either a premise or a claim.

Natural language processing methods identify, retrieve, generate, or select
omitted argument components, especially premises
\cite{Habernal.et.al.2018.NAACL.ARCT,singh-etal-2022-irac,
doi:10.1177/19462174251344764}, but generally do not expose the logical
inference completed by a selected or generated component.
Symbolic approaches make this inference explicit through abduction and other formal
reasoning methods
\cite{10.5555/1619645.1619657,10.1007/978-3-642-23963-2_13,
Black2012,Hosseini2014,Xydis2020,KR2022-27,Hunter_2022,
ORBELEIVA2023108963,Leiva2025,KR2025-11,ijcai2025p495}, but generally assume
that the formulae, candidate completions, or background knowledge are
available.

A prior neuro-symbolic pipeline combined candidate generation with logical
verification~\cite{Feng2026}. It generated two missing-premise candidates,
translated the stated and generated text into propositional formulae, and
tested each candidate independently with a SAT solver. Because each candidate
received only an entailment or non-entailment result, equal outcomes left the
pair unresolved. 

We instead assign each candidate a logical-resistance score and compare the
two scores.
Each candidate \(c\) defines a source statement collection \(S_c\) evaluated
against a target statement collection \(T_c\). Missing-premise tasks vary
\(S_c\) while holding \(T_c\) fixed; missing-claim tasks hold \(S_c\) fixed
while varying \(T_c\). Because the source and target are translated
separately, semantically corresponding or conflicting propositions remain
distinct Boolean atoms. We represent these cross-formula semantic relations
as \emph{atom links} encoding exact correspondence, entailment, or
contradiction.
This creates \emph{semantic-link uncertainty}: a target atom without an exact
correspondence may have several possible entailment or contradiction links,
as well as the option of remaining unlinked, and committing to one
configuration can change the candidate score.

We instantiate the pairwise logical-resistance score in two ways.
\textsc{Top-Link} serves as a single-configuration reference: for each such
target atom, it selects the highest-confidence entailment or contradiction
link when one is available and evaluates the resulting configuration with
weighted Partial MaxSAT. \emph{Possible-World Atom-Link Formalization} (PWAL)
instead keeps the translated statement formulae and exact atom correspondences
fixed, assigns a uniform distribution over each target atom's available
entailment or contradiction links and the option of leaving it unlinked, and
treats these local choices as independent. PWAL selects the candidate with
lower expected logical resistance. A candidate's logical-resistance score is
lower when requiring the target to hold creates less conflict and when more
target content is already supported without that requirement. The resulting
trace records the translated formulae, evaluated link configurations, and
resistance components used in each comparison.

We evaluate \textsc{Top-Link} and PWAL on five pairwise tasks. The Argument
Reasoning Comprehension Task
(ARCT)~\cite{Habernal.et.al.2018.NAACL.ARCT} and a task derived from
Context-Dependent Evidence Detection
(CDED)~\cite{rinott-etal-2015-show} evaluate missing-premise selection.
Tasks derived from iDebate~\cite{wang-ling-2016-neural} and Argument
Annotated Essays v2 (AAE2)~\cite{stab-gurevych-2017-parsing} evaluate
missing-claim selection. We also evaluate Abductive Natural Language
Inference (\(\alpha\)NLI)~\cite{DBLP:journals/corr/abs-1908-05739} as a
pairwise task for abductive hypothesis selection. The four argumentation tasks
cover both premise- and claim-side completion, while \(\alpha\)NLI evaluates
the same pairwise selection framework in an abductive setting.

\section{Background and Related Work}

\paragraph{Related Work}
Neuro-symbolic reasoning systems translate natural-language problems into
formal representations and invoke symbolic solvers
\cite{pan2023logic,olausson2023linc,ye2023satlm,kirtania-etal-2024-logic}.
Some refine a generated formalization using solver feedback or aggregate
predictions across several generated formalizations
\cite{pan2023logic,olausson2023linc}. Recent enthymeme methods compare supplied
premise candidates through natural-language multi-agent
debate~\cite{ku-etal-2025-multi}, optimize logical formalizations over argument
maps~\cite{ijcai2025p495}, or axiomatize the evaluation of alternative
formalizations~\cite{KR2025-11}. PWAL instead keeps each statement formula
fixed and marginalizes over alternative cross-formula atom-link configurations
during pairwise selection.

Weighted and probabilistic logic frameworks attach weights to formulae or
rules
\cite{richardson2006markov,bach2017hinge,riegel2020logical}.
PWAL uses soft-clause weights for within-world violation costs and world
probabilities for uncertainty over which links are active.

\paragraph{Translation of AMR into Logic}
Abstract Meaning Representation (AMR) represents the semantic structure of a
sentence as a rooted, labeled, directed graph
\cite{banarescu-etal-2013-abstract}. Nodes denote concepts or PropBank frames.
In a frame label such as \texttt{chase-01}, \texttt{chase} is the predicate
lemma and \texttt{01} is its PropBank sense identifier. Labeled edges encode
PropBank-derived core roles such as \texttt{:ARG0} and \texttt{:ARG1}
\cite{kingsbury-palmer-2002-treebank}, alongside general AMR relations such as
\texttt{:location} and \texttt{:time}
\cite{banarescu-etal-2013-abstract}. AMR also represents coordination,
conditions, polarity, and reentrancy.

Motivated by compositional AMR semantics
\cite{bos-2016-squib}, we use a fixed rule-based compiler that maps an AMR
graph to structured semantic atoms and a propositional formula. Coordination,
conditions, and formula-level polarity introduce conjunction or disjunction,
implication, and negation, respectively, before truth-preserving Boolean
simplification.

\begin{definition}[Statement-Level Propositional Representation]
\label{def:statement_representation}

For a finite semantic-atom set \(\mathcal A\), let
\(\mathcal L(\mathcal A)\) be the smallest formula language containing every
\(a\in\mathcal A\) and closed under \(\neg\), \(\land\), \(\lor\), and
\(\to\).

For each successfully translated statement \(x\), the solver-facing
representation is
\(\mathsf{Rep}(x)=\langle\mathcal A_x,\Phi_x,\mathcal V_x\rangle\), where:
\begin{itemize}
    \item \(\mathcal A_x\) is the finite, nonempty set of active semantic atoms;
    \item \(\Phi_x\in\mathcal L(\mathcal A_x)\) is the propositional formula
          compiled from the AMR structure; and
    \item \(\mathcal V_x\) maps each atom in \(\mathcal A_x\) to a deterministic,
          nonempty surface verbalization ending in a full stop.
          Formula-level polarity is represented in \(\Phi_x\) rather than in
          \(\mathcal V_x\).
\end{itemize}

Let \(\operatorname{Atoms}(\Phi_x)\) denote the atoms occurring in \(\Phi_x\).
The translator satisfies
\(\operatorname{Atoms}(\Phi_x)=\mathcal A_x\), so the atom inventory is
exactly the set of propositional variables occurring in the formula.
\end{definition}

The main atom forms are
\(\operatorname{Uni}(u)\), \(\operatorname{Dya}_{\kappa}(u,v)\), and
\(\operatorname{Tri}_{\kappa_a,\kappa_b}(u,e,v)\). Here \(u,v,e\) are AMR terms,
with \(e\) denoting the predicate occurrence shared by the two roles in a
triple, and \(\kappa,\kappa_a,\kappa_b\) denoting AMR roles. A unary atom retains a concept
that would otherwise have no active proposition or serves as a carrier for
local negation. A dyadic atom represents one role-labeled semantic relation.
A triple combines two roles licensed by the same predicate occurrence. Additional atom types, scope rules, and fallback cases are
specified in the supplementary material.

\begin{example}[AMR-to-Logic Translation]
\label{ex:amr_translation}

Consider \(x^+\), \emph{A cute dog chases a cat}, and the corresponding
negative statement \(x^-\), \emph{No cute dog chases a cat}.
After omitting token alignments and renaming variables, their parsed AMRs
differ only in the root polarity attribute:

\smallskip
\noindent
\begin{minipage}[t]{0.48\linewidth}
\vspace{0pt}
\centering
\textit{Positive AMR}\\
\raggedright
\texttt{(c / chase-01}\\
\quad\texttt{:ARG0 (d / dog}\\
\qquad\texttt{:mod (u / cute))}\\
\quad\texttt{:ARG1 (a / cat))}
\end{minipage}
\hfill
\begin{minipage}[t]{0.48\linewidth}
\vspace{0pt}
\centering
\textit{Negative AMR}\\
\raggedright
\texttt{(c / chase-01}\\
\quad\texttt{:polarity -}\\
\quad\texttt{:ARG0 (d / dog}\\
\qquad\texttt{:mod (u / cute))}\\
\quad\texttt{:ARG1 (a / cat))}
\end{minipage}
\par\smallskip

Using the same names for corresponding atoms, the modifier produces
\(a_1=\operatorname{Dya}_{\mathrm{mod}}
(\textit{dog},\textit{cute})\), while the shared
\texttt{:ARG0} and \texttt{:ARG1} roles produce
\(a_2=\operatorname{Tri}_{\mathrm{ARG0},\mathrm{ARG1}}
(\textit{dog},\textit{chase-01},\textit{cat})\).
Both translations have
\(\mathcal A_{x^+}=\mathcal A_{x^-}=\{a_1,a_2\}\) and the same surface
verbalizations, \textit{``cute dog.''} and
\textit{``dog chase cat.''} The sense identifier remains part of the
structured atom identity but is omitted from its surface verbalization.
Polarity changes the propositional formula:
\(\Phi_{x^+}=a_2\land a_1\), whereas
\(\Phi_{x^-}=\neg(a_2\land a_1)\).
\end{example}

\section{Pipeline}
\label{section:pipeline}

A pairwise instance contains two candidate-specific source--target pairs,
\((S_A,T_A)\) and \((S_B,T_B)\). For \(c\in\{A,B\}\), let
\(S_c=(s_{c,1},\ldots,s_{c,n_c})\) and
\(T_c=(t_{c,1},\ldots,t_{c,m_c})\) denote the source and target statement
collections for candidate \(c\). Statements within each collection are
conjoined. Common-source tasks satisfy \(S_A=S_B\), while common-target tasks
satisfy \(T_A=T_B\).

The pipeline translates and assembles each candidate-specific source--target
pair, constructs cross-formula atom links, and computes candidate-level
logical resistance using either \textsc{Top-Link} or PWAL; it selects the
lower-scoring candidate or returns a tie.

For each resistance-based scoring method, let \(\sigma_c\) denote the score
assigned to candidate \(c\).
Lower logical scores are preferred. The pipeline returns \texttt{Tie} when
\(\sigma_A=\sigma_B\); otherwise, it selects the candidate with the lower
score. This compares the two candidate-specific evaluations rather than
assigning either candidate an absolute validity label.

\subsection{Candidate Source and Target Assembly}
\label{section:source_target_assembly}

Each statement in \(S_c\) and \(T_c\) is translated independently. Before
assembly, statement-local atoms are renamed so that atom namespaces are
disjoint across all statement occurrences in \(S_c\) and \(T_c\). The same
renaming is applied to each statement's formula and verbalization map. Below,
\(\mathcal A_{x_i}\), \(\Phi_{x_i}\), and \(\mathcal V_{x_i}\) denote these
renamed representations.

\begin{definition}[Candidate Source and Target Representation]
\label{def:source_target_representation}

For each nonempty collection \(X\in\{S_c,T_c\}\), write
\(X=(x_1,\ldots,x_{|X|})\), and define
\(\mathcal A_X=\biguplus_{i=1}^{|X|}\mathcal A_{x_i}\) and
\(\Phi_X=\bigwedge_{i=1}^{|X|}\Phi_{x_i}\), where \(\biguplus\) denotes
disjoint union. For each \(a\in\mathcal A_X\), let \(x_i\) be its unique
originating statement and set
\(\mathcal V_X(a)=\mathcal V_{x_i}(a)\). The source and target namespaces
satisfy \(\mathcal A_{S_c}\cap\mathcal A_{T_c}=\varnothing\).

Let \(\operatorname{CNF}(\Phi)\) denote the clause collection in conjunctive
normal form (CNF) produced by the fixed SymPy-based conversion used by the solver
\cite{10.7717/peerj-cs.103}. For \(X\in\{S_c,T_c\}\), define
\(\Gamma_X=\biguplus_{i=1}^{|X|}\operatorname{CNF}(\Phi_{x_i})\).
For clauses, \(\biguplus\) preserves statement-indexed clause occurrences.
A literal is an atom or its negation; a clause is a disjunction of literals,
written \([\ell_1\lor\cdots\lor\ell_m]\). Then
\(\Phi_X\equiv\bigwedge_{q\in\Gamma_X}q\).
\end{definition}

The fixed CNF conversion introduces no auxiliary Boolean atoms.

\begin{example}[Source and Target Assembly]
\label{ex:source_target_assembly}

Suppose that, after renaming,
\(\Phi_{s_{c,1}}=a_0\),
\(\Phi_{s_{c,2}}=a_1\),
\(\Phi_{s_{c,3}}=\neg(a_2\land a_3)\), and
\(\Phi_{t_{c,1}}=b_1\to b_2\).
Then
\(\Phi_{S_c}=a_0\land a_1\land\neg(a_2\land a_3)\) and
\(\Phi_{T_c}=b_1\to b_2\), with
\(\Gamma_{S_c}=\{[a_0],[a_1],[\neg a_2\lor\neg a_3]\}\) and
\(\Gamma_{T_c}=\{[\neg b_1\lor b_2]\}\).
\end{example}

\subsection{Candidate Atom Links}
\label{section:candidate_links}

Because source and target atoms are distinct Boolean variables, their semantic
relations must be represented explicitly. An \emph{atom link} connects a
source atom to a target atom.

\begin{definition}[Natural Language Inference]
\label{def:atom_nli}

Let
\(\mathcal Y=\{\texttt{Ent},\texttt{Con},\texttt{Neu}\}\)
denote the entailment, contradiction, and neutral labels. For nonempty verbalization
strings \(u\) and \(v\), let
\(\mathsf N(u,v)=(y,p)\in\mathcal Y\times[0,1]\), where \(u\) is the NLI
premise, \(v\) is the hypothesis, \(y\) is the predicted label, and \(p\) is
the confidence assigned to \(y\), rounded to three decimal places.
\end{definition}

\begin{definition}[Exact and NLI-Derived Atom Links]
\label{def:candidate_atom_links}
Let \(\operatorname{cf}(u)\) denote the case-folded form of string \(u\).
Define
\[
\begin{aligned}
\mathcal M_c^{\mathrm{ex}}
&=
\{(a,b)\in\mathcal A_{S_c}\times\mathcal A_{T_c}
  \mid \operatorname{cf}(\mathcal V_{S_c}(a))
  =\operatorname{cf}(\mathcal V_{T_c}(b))\},\\
\mathcal U_c
&=
\{b\in\mathcal A_{T_c}
  \mid \nexists a\in\mathcal A_{S_c}:
  (a,b)\in\mathcal M_c^{\mathrm{ex}}\}.
\end{aligned}
\]
Every matching source--target pair is retained, so a target atom may have
several exact links. If a target atom has an exact link, it is not passed to
NLI.

For each \(a\in\mathcal A_{S_c}\) and \(b\in\mathcal U_c\), let
\((y_{a,b},p_{a,b})
=\mathsf N(\mathcal V_{S_c}(a),\mathcal V_{T_c}(b))\).
The retained non-neutral alternatives are
\(\mathcal R_c(b)=\{(a,b,y_{a,b},p_{a,b})\mid
a\in\mathcal A_{S_c},\ y_{a,b}\in\{\texttt{Ent},\texttt{Con}\}\}\).
Neutral outputs generate no link. \(\mathcal R_c(b)\) contains the alternative
non-exact links for the unmatched target atom \(b\).
\end{definition}

\begin{example}[Exact and NLI-Derived Links]
\label{ex:candidate_atom_links}

Suppose
\(\mathcal V_{S_c}(a_1)=\mathcal V_{S_c}(a_2)
=\mathcal V_{T_c}(b_1)=\textit{``dog bark.''}\),
and these are the only exact matches. Then
\(\mathcal M_c^{\mathrm{ex}}=\{(a_1,b_1),(a_2,b_1)\}\).

Let \(b_2\) have verbalization \textit{``animal move.''} and no exact match.
If the NLI outputs for the pairs \((a_3,b_2)\) and \((a_4,b_2)\) are
\((\texttt{Ent},0.88)\) and \((\texttt{Con},0.83)\), respectively, and all
other comparisons with \(b_2\) are neutral, then
\(\mathcal R_c(b_2)
=\{(a_3,b_2,\texttt{Ent},0.88),
(a_4,b_2,\texttt{Con},0.83)\}\).
\end{example}
\subsection{Weighted Atom-Link Clauses}
\label{section:weighted_links}

Atom links are encoded as weighted clauses. We write \((q,\infty)\) for a hard
clause and \((q,w)\), with \(w\in\mathbb N_{>0}\), for a soft clause. Every
hard clause must be satisfied, while \(w\) is the penalty incurred when the
soft clause \(q\) is violated.

\begin{definition}[Weighted Atom-Link Clauses]
\label{def:weighted_link_clauses}

Let \(W_{\max}\in\mathbb N_{>0}\) be the maximum finite link weight and define
\(w(p)=\max\{1,\lfloor W_{\max}p\rfloor\}\) for \(p\in[0,1]\), where
\(\lfloor x\rfloor\) is the greatest integer not exceeding \(x\). Thus,
\(w(p)\in\{1,\ldots,W_{\max}\}\), with the outer maximum ensuring a positive
weight.

Each exact link \((a,b)\in\mathcal M_c^{\mathrm{ex}}\) contributes
\(([\neg a\lor b],W_{\max})\) and
\(([a\lor\neg b],W_{\max})\), encoding \(a\rightarrow b\) and
\(b\rightarrow a\), respectively. Together they form a soft equivalence.
Define \(\Omega_c^{\mathrm{ex}}
=\bigcup_{(a,b)\in\mathcal M_c^{\mathrm{ex}}}
\{([\neg a\lor b],W_{\max}),([a\lor\neg b],W_{\max})\}\).

For a non-exact link \(r=(a,b,y,p)\), let
\(\operatorname{cl}(r)=([\neg a\lor b],w(p))\) if
\(y=\texttt{Ent}\), and
\(\operatorname{cl}(r)=([\neg a\lor\neg b],w(p))\) if
\(y=\texttt{Con}\).

An active non-exact link set
\(L_c\subseteq\bigcup_{b\in\mathcal U_c}\mathcal R_c(b)\)
is admissible if
\(|L_c\cap\mathcal R_c(b)|\leq1\) for every
\(b\in\mathcal U_c\). Thus, each unmatched target atom has at most one active
non-exact link. The atom-link clause collection induced by \(L_c\) is
\(\Omega_c^{\mathrm{sem}}(L_c)
=\Omega_c^{\mathrm{ex}}
\cup\{\operatorname{cl}(r)\mid r\in L_c\}\).
\end{definition}
Exact-link clauses occur in \(\Omega_c^{\mathrm{sem}}(L_c)\) for every
admissible \(L_c\).

\begin{example}[Weighted Atom-Link Clauses]
\label{ex:weighted_link_clauses}

Continuing Example~\ref{ex:candidate_atom_links}, let \(W_{\max}=100\).
For each \(i\in\{1,2\}\), the exact link \((a_i,b_1)\) contributes
\(([\neg a_i\lor b_1],100)\) and
\(([a_i\lor\neg b_1],100)\).
The two non-exact alternatives map to
\(([\neg a_3\lor b_2],88)\) and
\(([\neg a_4\lor\neg b_2],83)\).
An admissible configuration may activate either non-exact link for \(b_2\),
or neither, but not both.
\end{example}

\subsection{Partial MaxSAT Evaluation}
\label{section:weighted_evaluator}

For a fixed active non-exact link set \(L_c\), the evaluator compares a base
instance with a forced instance that additionally requires the target clauses.
All hard clauses must be satisfied, and the solver minimizes the total weight
of violated soft clauses.

\begin{definition}[Base and Forced MaxSAT Instances]
\label{def:base_forced_instances}

The hard source clauses are
\(\Omega_c^{\mathrm{hard}}
=\{(q,\infty)\mid q\in\Gamma_{S_c}\}\), and the target-inertia clauses are
\(\Omega_c^{\mathrm{inertia}}
=\{([\neg b],\epsilon)\mid b\in\mathcal A_{T_c}\}\), where
\(\epsilon\in\mathbb N_{>0}\). Inertia prefers false target atoms but does not
require them. Define the base and forced instances by
\[
\begin{aligned}
\mathcal I_c^{\mathrm{base}}(L_c)
&=\Omega_c^{\mathrm{hard}}\cup\Omega_c^{\mathrm{sem}}(L_c)
\cup\Omega_c^{\mathrm{inertia}},\\
\mathcal I_c^{\mathrm{forced}}(L_c)
&=\mathcal I_c^{\mathrm{base}}(L_c)
\cup\{(q,\infty)\mid q\in\Gamma_{T_c}\}.
\end{aligned}
\]

Let \(M_c^{\mathrm{base}}(L_c)\) and
\(M_c^{\mathrm{forced}}(L_c)\) be the assignments returned by solving
\(\mathcal I_c^{\mathrm{base}}(L_c)\) and
\(\mathcal I_c^{\mathrm{forced}}(L_c)\), respectively. Each assignment
satisfies all hard clauses and minimizes the total weight of violated soft
clauses. When either instance has multiple optimal assignments, the
corresponding \(M_c^{\mathrm{base}}\) or
\(M_c^{\mathrm{forced}}\) is the assignment returned by the solver; no
secondary optimization criterion is applied. For any assignment \(M\), define
\[
\operatorname{SemCost}_c(M,L_c)
=
\sum_{\substack{(q,w)\in\Omega_c^{\mathrm{sem}}(L_c)\\
M\not\models q}}
w.
\]
Both assignments minimize atom-link and inertia costs jointly. Inertia
provides a conservative default for otherwise unconstrained target atoms; as
an optimization regularizer rather than atom-link evidence, it is excluded
from \(\operatorname{SemCost}_c\).

The \emph{semantic tension} is the signed change
\[
\begin{aligned}
\Delta T_c(L_c)
&=\operatorname{SemCost}_c(M_c^{\mathrm{forced}}(L_c),L_c)\\
&\quad-\operatorname{SemCost}_c(M_c^{\mathrm{base}}(L_c),L_c).
\end{aligned}
\]
\end{definition}

A positive \(\Delta T_c(L_c)\) means that enforcing the target increases
conflict with the active atom links; a negative value means that it
reduces this conflict.

Semantic tension does not indicate which target clauses already hold in the
base optimum.
We therefore define a target-clause witness ratio
\(R_{\mathrm{sat},c}(L_c)\). The condition
\(\operatorname{NegSup}_{c,L_c}(b)\) prevents target inertia alone from
witnessing a negative target literal. Logical resistance combines normalized
semantic tension with this ratio. A lower value reflects lower normalized
semantic tension, stronger target support, or both; a higher value reflects
greater tension, weaker support, or both.

\begin{definition}[Target Witness and Logical Resistance]
\label{def:logical_resistance}

Write \(M_c^0=M_c^{\mathrm{base}}(L_c)\). For a target atom \(b\), define
\(\operatorname{NegSup}_{c,L_c}(b)\) to hold if either there exists
\(a\in\mathcal A_{S_c}\) such that
\((a,b)\in\mathcal M_c^{\mathrm{ex}}\) and \(M_c^0\models\neg a\), or there
exists \((a,b,\texttt{Con},p)\in L_c\) such that \(M_c^0\models a\).
We refer to this requirement as the \emph{polarity-aware negative guard}.

For a target literal \(\ell\) over atom \(b\), let
\(\operatorname{Wit}_{c,L_c}(\ell)=1\) if either
\(\ell=b\) and \(M_c^0\models b\), or
\(\ell=\neg b\), \(M_c^0\models\neg b\), and
\(\operatorname{NegSup}_{c,L_c}(b)\); otherwise it is \(0\).
For \(q\in\Gamma_{T_c}\), let
\(\operatorname{Wit}_{c,L_c}(q)
=\max_{\ell\in q}\operatorname{Wit}_{c,L_c}(\ell)\).
For \(\Gamma_{T_c}\neq\varnothing\), the target-clause witness ratio is
\[
R_{\mathrm{sat},c}(L_c)
=|\Gamma_{T_c}|^{-1}
\sum_{q\in\Gamma_{T_c}}\operatorname{Wit}_{c,L_c}(q).
\]
Because the witness ratio is clause-based, the fixed CNF conversion keeps
clause granularity consistent across scoring methods.

The total included link weight is
\[
W_c(L_c)=W_{\max}|\mathcal M_c^{\mathrm{ex}}|
+\sum_{(a,b,y,p)\in L_c}w(p).
\]
Each exact link contributes \(W_{\max}\), because an assignment can violate
at most one of its two equivalence clauses. Let
\[
C_{\max,c}(L_c)=
\begin{cases}
W_c(L_c), & W_c(L_c)>0,\\
|\mathcal A_{T_c}|W_{\max}, & \text{otherwise}.
\end{cases}
\]
We use \(C_{\max,c}(L_c)\) as the normalization capacity; it is an upper bound
on the active atom-link cost but need not be attainable. When \(W_c(L_c)=0\), no
atom-link clause is active and
\(\Delta T_c(L_c)=0\), so the fallback denominator only prevents division by
zero. The logical resistance is
\[
\rho_c(L_c)
=\frac{\Delta T_c(L_c)}{C_{\max,c}(L_c)}
-R_{\mathrm{sat},c}(L_c).
\]
\end{definition}

Both resistance components are computed from solver-returned optima. In
particular, the witness ratio is not logical entailment over all assignments
satisfying the source clauses.

\begin{example}[Logical Resistance]
\label{ex:logical_resistance}

Let \(\Gamma_{S_c}=\{[a]\}\), \(\Gamma_{T_c}=\{[b]\}\),
\(W_{\max}=100\), and \(\epsilon=1\), with no exact links.
For \(r_E=(a,b,\texttt{Ent},0.80)\) and \(L_c=\{r_E\}\), the hard source
clause forces \(a\) true. The base optimum sets \(b\) true because violating
inertia costs \(1\), whereas violating the entailment link costs \(80\); the
forced optimum also sets \(b\) true. Both atom-link costs are \(0\), so
\(\Delta T_c(L_c)=0\), \(R_{\mathrm{sat},c}(L_c)=1\),
\(C_{\max,c}(L_c)=80\), and \(\rho_c(L_c)=-1\).

For \(r_C=(a,b,\texttt{Con},0.70)\) and \(L_c=\{r_C\}\), the base optimum
sets \(b\) false, while the forced optimum sets \(b\) true and violates the
weight-\(70\) contradiction link. Although the forced optimum also violates
inertia, inertia is excluded from \(\operatorname{SemCost}_c\). Thus
\(\Delta T_c(L_c)=70\), \(R_{\mathrm{sat},c}(L_c)=0\),
\(C_{\max,c}(L_c)=70\), and \(\rho_c(L_c)=1\).

If the target instead contains \(\neg b\) without an exact or contradiction
link supporting that polarity, inertia may set \(b\) false, but
\(\neg b\) is not witnessed.
\end{example}

\begin{figure*}[t]
\centering
\includegraphics{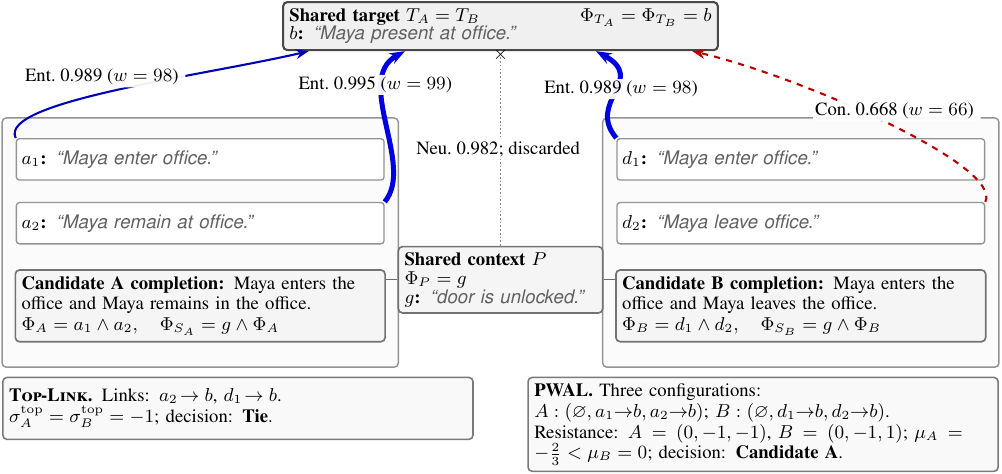}

\caption{Constructed common-target trace. Atom identifiers precede their
italic sans-serif verbalizations. Arrows show confidence and clause weight
(blue: entailment; dashed red: contradiction; thick:
\textsc{Top-Link}). PWAL treats non-neutral links and no-link as local
alternatives; dotted neutral is discarded.}
\label{fig:pwal_trace}
\end{figure*}

\subsection{\textsc{Top-Link} and PWAL}
\label{section:top_link_pwal}

The two scoring methods use the same source and target formulae, exact links,
non-exact link inventory, clause-construction rule, and MaxSAT evaluator.
\textsc{Top-Link} evaluates one admissible non-exact link configuration,
whereas PWAL averages logical resistance over a distribution of such
configurations.

\begin{definition}[\textsc{Top-Link}]
\label{def:top_link}

For each \(b\in\mathcal U_c\) with
\(\mathcal R_c(b)\neq\varnothing\), define
\(r_c^\star(b)\in
\arg\max_{(a,b,y,p)\in\mathcal R_c(b)}p\)
as a maximum-confidence link, breaking confidence ties by a fixed total
order on \(\mathcal A_{S_c}\) induced by source-statement
order and the translator's atom order. Set
\(L_c^{\mathrm{top}}=\{r_c^\star(b)\mid b\in\mathcal U_c,\,
\mathcal R_c(b)\neq\varnothing\}\) and
\(\sigma_c^{\mathrm{top}}=\rho_c(L_c^{\mathrm{top}})\).
If \(\mathcal R_c(b)=\varnothing\), no non-exact link is selected for \(b\).
\end{definition}

PWAL treats every non-neutral link and the no-link outcome as local
alternatives for each unmatched target atom.

\begin{definition}[PWAL]
\label{def:pwal_worlds}

For each \(b\in\mathcal U_c\), define
\(\mathcal Z_{c,b}=\{\varnothing\}\cup\mathcal R_c(b)\), where
\(\varnothing\) denotes no active non-exact link. PWAL assigns every local
outcome equal probability:
\(\pi_{c,b}(z)=1/(|\mathcal R_c(b)|+1)\) for
\(z\in\mathcal Z_{c,b}\). If
\(\mathcal R_c(b)=\varnothing\), then
\(\pi_{c,b}(\varnothing)=1\).

Let
\(\mathcal W_c=\prod_{b\in\mathcal U_c}\mathcal Z_{c,b}\).
Each \(\omega_c=(z_b)_{b\in\mathcal U_c}\in\mathcal W_c\) activates
\(L_c^{\omega_c}=\{z_b\mid z_b\neq\varnothing\}\).
Assuming target-wise independent local link choices,
\(\pi_c(\omega_c)=\prod_{b\in\mathcal U_c}\pi_{c,b}(z_b)
=1/N_c^{\mathrm{world}}\), where
\(N_c^{\mathrm{world}}=|\mathcal W_c|
=\prod_{b\in\mathcal U_c}(|\mathcal R_c(b)|+1)\).
Logical dependencies among target atoms remain encoded in \(\Phi_{T_c}\) and
the MaxSAT instances. Exact-link clauses are included in every world and add
no local choices.

When all worlds are enumerated, the candidate score is
\[
\sigma_c^{\mathrm{PWAL}}=\mu_c
=\sum_{\omega_c\in\mathcal W_c}
\pi_c(\omega_c)\rho_c(L_c^{\omega_c}).
\]
For \(K\in\mathbb N_{>0}\) i.i.d.\ sampled worlds
\(\omega_c^{(1)},\ldots,\omega_c^{(K)}\sim\pi_c\), it is estimated by
\[
\widehat\sigma_c^{\mathrm{PWAL}}=\widehat\mu_c
=K^{-1}\sum_{k=1}^{K}\rho_c(L_c^{\omega_c^{(k)}}).
\]
Thus, \(\widehat\mu_c\) is a finite-\(K\) Monte Carlo estimator of
\(\mu_c\), the expected logical resistance.
\end{definition}

NLI confidence determines the within-world clause weight \(w(p)\), not the
world probability. A world specifies one alternative cross-formula atom-link
configuration: the statement formulae and exact-link clauses remain fixed,
while the active non-exact links vary. PWAL compares expected resistance
rather than the number of worlds won by each candidate.

Figure~\ref{fig:pwal_trace} gives a trace-level illustration of this
mechanism. 

\begin{table*}[t]
\centering
\begin{tabular}{@{}llrrrrrr@{}}
\toprule
Dataset (\(N\)) & Metric & \textsc{FH-HardSAT} & \textsc{Top-Link} & PWAL
& \shortstack{PWAL\\w/o no-link}
& \shortstack{PWAL\\w/o neg.\ guard}
& Direct NLI \\
\midrule
\multirow{3}{*}{\(\alpha\)NLI (400)}
& Acc  & 11.00 & 36.50 & \(\mathbf{49.55\pm0.96}\) & \(46.15\pm0.41\) & \(49.48\pm0.89\) & 57.75 \\
& Tie  & 55.75 & 27.25 & \(2.15\pm0.50\) & \(11.98\pm0.14\) & \(4.63\pm0.38\) & 0.00 \\
& EAcc & 38.88 & 50.13 & \(50.63\pm0.98\) & \(52.14\pm0.41\) & \(51.79\pm0.93\) & 57.75 \\
\midrule
\multirow{3}{*}{ARCT (444)}
& Acc  & 7.43 & 20.05 & \(\mathbf{50.90\pm1.12}\) & \(47.77\pm1.23\) & \(40.59\pm1.12\) & 70.05 \\
& Tie  & 86.71 & 64.86 & \(6.87\pm0.41\) & \(17.32\pm0.20\) & \(31.89\pm0.28\) & 0.00 \\
& EAcc & 50.79 & 52.48 & \(54.34\pm1.07\) & \(56.43\pm1.17\) & \(56.53\pm1.10\) & 70.05 \\
\midrule
\multirow{3}{*}{CDED (400)}
& Acc  & 34.50 & 62.00 & \(\mathbf{73.50\pm0.79}\) & \(71.50\pm0.50\) & \(69.60\pm0.53\) & 80.00 \\
& Tie  & 49.75 & 13.50 & \(2.58\pm0.29\) & \(5.58\pm0.21\) & \(8.23\pm0.25\) & 0.00 \\
& EAcc & 59.38 & 68.75 & \(74.79\pm0.73\) & \(74.29\pm0.53\) & \(73.71\pm0.51\) & 80.00 \\
\midrule
\multirow{3}{*}{iDebate (400)}
& Acc  & 33.00 & 60.75 & \(\mathbf{63.70\pm0.65}\) & \(63.48\pm0.56\) & \(63.63\pm0.53\) & 72.50 \\
& Tie  & 55.00 & 5.00 & \(0.00\pm0.00\) & \(0.03\pm0.08\) & \(0.33\pm0.12\) & 0.00 \\
& EAcc & 60.50 & 63.25 & \(63.70\pm0.65\) & \(63.49\pm0.56\) & \(63.79\pm0.56\) & 72.50 \\
\midrule
\multirow{3}{*}{AAE2 (350)}
& Acc  & 18.86 & 51.14 & \(\mathbf{54.43\pm0.56}\) & \(53.89\pm0.45\) & \(54.17\pm0.72\) & 70.57 \\
& Tie  & 62.00 & 7.43 & \(2.86\pm0.00\) & \(3.74\pm0.09\) & \(2.37\pm0.14\) & 0.00 \\
& EAcc & 49.86 & 54.86 & \(55.86\pm0.56\) & \(55.76\pm0.44\) & \(55.36\pm0.69\) & 70.57 \\
\bottomrule
\end{tabular}
\caption{Pairwise test results (\%). PWAL variants report ten-seed means
\(\pm\) sample SD (\(K=100\)); other methods are deterministic. Failed
\textsc{FH-HardSAT} evaluations count as errors; bold marks the best Acc among
non-ablated logical methods.}
\label{tab:main_results}
\end{table*}

\section{Experimental Setup}
\label{section:experimental_setup}

\paragraph{Tasks and pair construction.}
Each instance contains two candidate-specific source--target pairs,
\((S_A,T_A)\) and \((S_B,T_B)\), with one designated gold candidate.
We use ARCT~\cite{Habernal.et.al.2018.NAACL.ARCT} and
\(\alpha\)NLI~\cite{DBLP:journals/corr/abs-1908-05739} in their original
pairwise form, and derive pairwise tasks from
CDED~\cite{rinott-etal-2015-show},
iDebate~\cite{wang-ling-2016-neural}, and
AAE2~\cite{stab-gurevych-2017-parsing}.

ARCT uses \(S_c=(\text{reason},\text{warrant}_c)\) and
\(T_c=(\text{claim})\). \(\alpha\)NLI uses
\(S_c=(O_1,\text{hypothesis}_c)\) and \(T_c=(O_2)\).

For the derived tasks, superscripts \(+\) and \(-\) denote the gold and
distractor items, not candidate positions or stance labels. In CDED,
\(T_A=T_B=(C_i)\) and
\(\{S_A,S_B\}=\{(H_i^+),(H_i^-)\}\), where \(H_i^+\) is annotated as
evidence for \(C_i\), and \(H_i^-\) is a same-topic passage annotated as
evidence for another claim but not annotated as evidence for \(C_i\).

For the two derived missing-claim tasks, let
\(P_i=(p_{i,1},\ldots,p_{i,n_i})\) denote the source statement collection
paired with gold claim \(C_i^+\). We set \(S_A=S_B=P_i\) and
\(\{T_A,T_B\}=\{(C_i^+),(C_i^-)\}\). In iDebate, \(P_i\) is the
argumentative statement collection associated with central claim \(C_i^+\),
and \(C_i^-\) is a different central claim from the same debate. In AAE2,
\(P_i\) contains all premise components directly annotated as supporting
\(C_i^+\), and \(C_i^-\) is an opposite-stance claim from the same essay
such that no premise in \(P_i\) has an annotated relation path to it. Because
stance is defined relative to the essay's major claim, \(C_i^-\) is not
assumed to be the logical negation of \(C_i^+\). 

We evaluate the full \(444\)-example ARCT test set, a fixed
\(400\)-instance sample from the official \(\alpha\)NLI test split, fixed
\(400\)-instance samples from the derived CDED and iDebate pools, and a fixed
\(350\)-instance AAE2 test set. Construction metadata are excluded from method
inputs; development and test groups are disjoint (see supplementary material).

\paragraph{Compared methods.}
The controlled comparison is between PWAL and \textsc{Top-Link}. Both use the
same source and target formulae, exact links, NLI-derived alternatives, clause
weights, and Partial MaxSAT evaluator. We additionally report
\textsc{FH-HardSAT}, the prior neuro-symbolic
method~\cite{Feng2026}, and \textsc{Direct NLI}.

\textsc{FH-HardSAT} evaluates the candidates independently and selects a
candidate only when exactly one is entailed. Equal valid outcomes produce a
tie. Its similarity and contradiction thresholds are selected separately for
each dataset using held-out development data.

\textsc{Direct NLI} uses \(u_c=\operatorname{cat}(S_c)\) and
\(v_c=\operatorname{cat}(T_c)\), where \(\operatorname{cat}(X)\) concatenates
the statements in \(X\) in their listed order, separated by a single space.
It applies the NLI model used for atom-link construction. Let \(p_\ell(u,v)\)
denote the probability assigned by this model to label \(\ell\in\mathcal Y\).
It scores candidate \(c\) by
\(s_c^{\mathrm{NLI}}=p_{\mathrm{Ent}}(u_c,v_c)-
p_{\mathrm{Con}}(u_c,v_c)\). It selects the candidate with the higher score;
score differences within the numerical tolerance specified below are treated
as ties. No such ties occur in the reported test sets.
It uses no AMR translation, atom links, or MaxSAT inference.

\paragraph{Implementation.}
\textsc{Top-Link}, PWAL, and the component ablations use the logical-resistance
score defined above, without reweighting its two components. We set
\(W_{\max}=100\), the target-inertia weight to \(\epsilon=1\), and the main
PWAL sampling budget to \(K=100\).
Sampled PWAL results, including ablations and each sampling budget, use seeds
2026--2035 and are reported as means \(\pm\) sample standard deviations.
Dataset-level metrics are computed separately for each seed before
aggregation. \textsc{Top-Link}, \textsc{FH-HardSAT}, \textsc{Direct NLI},
and the fixed exact/\(10^5\)-capped references are reported as single values.

We use Structured-BART for AMR parsing
\cite{zhou2021structure,lee2022maximum}, the frozen rule-based AMR-to-logic
compiler specified in the supplementary material, mDeBERTa-v3 for NLI-based
atom-link construction~\cite{he2023debertav3,laurer2024mdebertaxnli}, and RC2
in PySAT for Partial MaxSAT inference~\cite{Ignatiev2018}. In implementation,
\(\lvert\sigma_A-\sigma_B\rvert\leq\tau_{\mathrm{tie}}\) is treated as
a tie, with \(\tau_{\mathrm{tie}}=10^{-12}\).

\paragraph{Metrics.}
Let \(N_{\mathrm{win}},N_{\mathrm{err}},N_{\mathrm{tie}}\) denote correct
unique selections, errors, and valid ties, respectively, with
\(N=N_{\mathrm{win}}+N_{\mathrm{err}}+N_{\mathrm{tie}}\). A valid tie is a
successful evaluation returned as \texttt{Tie} by this decision rule. Errors
include incorrect unique selections and failed evaluations. We report
\(\operatorname{Accuracy}=N_{\mathrm{win}}/N\),
\(\operatorname{TieRate}=N_{\mathrm{tie}}/N\), and
\(\operatorname{EAcc}=(N_{\mathrm{win}}+\tfrac12N_{\mathrm{tie}})/N\).
Accuracy is the primary metric. EAcc gives valid ties half credit and errors
none, corresponding to uniform random tie-breaking for evaluation only.
We use pointwise 95\% paired two-way bootstrap intervals over
dataset-specific source clusters and seed runs (10,000 replicates).

\begin{table}[t]
\centering
\begin{tabular}{@{}llccc@{}}
\toprule
Dataset & Metric & \(K=100\) & \(K=200\) & Ref. \\
\midrule
\multirow{3}{*}{\(\alpha\)NLI}
& Acc  & \(49.55\pm0.96\) & \(49.92\pm1.16\) & \(49.75\) \\
& Tie  & \(2.15\pm0.50\) & \(1.48\pm0.25\) & \(3.25\) \\
& EAcc & \(50.63\pm0.98\) & \(50.66\pm1.17\) & \(51.38\) \\
\midrule
\multirow{3}{*}{ARCT}
& Acc  & \(50.90\pm1.12\) & \(52.00\pm1.59\) & \(48.20\) \\
& Tie  & \(6.87\pm0.41\) & \(6.71\pm0.51\) & \(14.41\) \\
& EAcc & \(54.34\pm1.07\) & \(55.36\pm1.47\) & \(55.41\) \\
\midrule
\multirow{3}{*}{CDED}
& Acc  & \(73.50\pm0.79\) & \(73.55\pm0.45\) & \(74.25\) \\
& Tie  & \(2.58\pm0.29\) & \(2.28\pm0.30\) & \(1.50\) \\
& EAcc & \(74.79\pm0.73\) & \(74.69\pm0.47\) & \(75.00\) \\
\midrule
\multirow{3}{*}{iDebate}
& Acc  & \(63.70\pm0.65\) & \(63.50\pm0.31\) & \(63.25\) \\
& Tie  & \(0.00\pm0.00\) & \(0.00\pm0.00\) & \(0.00\) \\
& EAcc & \(63.70\pm0.65\) & \(63.50\pm0.31\) & \(63.25\) \\
\midrule
\multirow{3}{*}{AAE2}
& Acc  & \(54.43\pm0.56\) & \(54.43\pm0.49\) & \(54.29\) \\
& Tie  & \(2.86\pm0.00\) & \(2.89\pm0.09\) & \(2.86\) \\
& EAcc & \(55.86\pm0.56\) & \(55.87\pm0.49\) & \(55.71\) \\
\bottomrule
\end{tabular}
\caption{Sampling and exact/\(10^5\)-capped reference results (\%).
Ref. denotes the exact/\(10^5\)-capped reference.}
\label{tab:sampling_exact}
\end{table}

\section{Results}
\label{section:results}

\paragraph{Main Results}

Table~\ref{tab:main_results} reports the test results. All differences
discussed below are computed from unrounded values.

\textsc{FH-HardSAT} returns valid ties on \(49.75\)--\(86.71\%\) of examples
and has lower Acc and EAcc than \textsc{Top-Link} and PWAL on every task.
Relative to \textsc{Top-Link}, PWAL raises mean Acc by
\(2.95\)--\(30.86\) percentage points, lowers mean Tie by
\(4.57\)--\(58.00\) percentage points, and raises mean EAcc by
\(0.45\)--\(6.04\) percentage points across the five tasks. On
\textsc{Top-Link} ties, PWAL's mean EAcc is
\(49.36\%\)--\(57.13\%\), versus \(50\%\) under uniform tie-breaking. Among
\textsc{Top-Link}'s unique decisions, mean
wrong-to-correct repairs exceed correct-to-wrong damages on all five tasks.
Full transition matrices are reported in the supplementary material.
Paired bootstrap analysis supports the Acc gains on \(\alpha\)NLI, ARCT, and
CDED, while the smaller Acc gains on iDebate and AAE2 remain uncertain.

\textsc{Direct NLI} has the highest EAcc on every task, exceeding PWAL by
\(5.21\)--\(15.71\) percentage points. It is a predictive reference rather
than a controlled logical comparator; PWAL and \textsc{Top-Link} share the
formalization and scoring pipeline except for the treatment of non-exact link
configurations.

Removing the no-link outcome lowers mean Acc and raises mean Tie on all five
tasks; the corresponding Full-minus-ablation Acc intervals exclude zero on
\(\alpha\)NLI, ARCT, and CDED. Removing the negative guard lowers mean Acc on
every task, with intervals excluding zero on ARCT and CDED. EAcc may increase
when an ablation produces additional ties because each valid tie receives
half credit.

\paragraph{Sampling Approximation and Exact/Capped Reference}

We evaluate \(K\in\{10,20,50,100,200\}\) over ten seeds. Within each seed,
the smaller budgets are prefixes of the same \(K=200\) world stream. For the
reference, a candidate is fully enumerated when its world count is at most
\(10^5\); otherwise, \(10^5\) worlds are sampled uniformly without replacement
using seed 2026. A pair is fully exact only when both candidate scores are
enumerated. The resulting fully exact pair counts are \(357/400\), \(443/444\),
\(216/400\), \(146/400\), and \(93/350\) for \(\alpha\)NLI, ARCT, CDED,
iDebate, and AAE2, respectively. Every other pair contains at least one capped
candidate. Table~\ref{tab:sampling_exact} reports \(K\in\{100,200\}\) and the
exact/\(10^5\)-capped reference; the full budget grid, timing protocol, and
runtime distributions are reported in the supplementary material.

At \(K=100\), the across-seed standard deviation is at most \(1.12\) points
for Acc, \(0.50\) for Tie, and \(1.07\) for EAcc. Increasing \(K\) from
\(100\) to \(200\) changes mean Acc by at most \(1.10\) points and mean Tie
by at most \(0.68\) points. Mean EAcc changes by at most \(0.20\) points on
four tasks and by \(1.02\) points on ARCT. The \(K=100\) mean EAcc is within
\(1.07\) points of the exact/\(10^5\)-capped reference on every task,
supporting \(K=100\) as a practical cost--stability setting.

ARCT has the largest sampled--reference difference. The reference raises Tie
from \(6.87\%\) to \(14.41\%\), lowers Acc from \(50.90\%\) to
\(48.20\%\), and raises EAcc from \(54.34\%\) to \(55.41\%\). All \(64\)
reference-tied ARCT pairs are fully exact, with absolute expected-resistance
margins of at most \(10^{-12}\). Finite-\(K\) Monte Carlo estimates can break
these ties, helping explain why sampled strict Acc is higher while sampled
EAcc is lower. On the other four tasks, reference EAcc differs from the
\(K=100\) mean by at most \(0.75\) points.

Restricting the analysis to fully exact pairs, \(K=100\) agrees with exact
marginalization on \(83.02\%\)--\(98.39\%\) of decisions across the five
tasks, averaged over ten seeds. Among \(K=100\)'s incorrect unique decisions
on these pairs, \(78.72\%\)--\(98.90\%\) remain incorrect under exact
marginalization. Thus, most of these errors are not removed by eliminating
finite-\(K\) approximation. Dataset-level analyses of score error, margin
error, decision transitions, and exact/capped strata are reported in the
supplementary material.

On the controlled fully enumerable runtime subset, \textsc{Top-Link} takes
\(1.10\)--\(1.68\) ms/example. Relative to \textsc{Top-Link}, the mean
task-level runtime multipliers are approximately \(84\times\), \(172\times\),
and \(5.8\times10^3\) for \(K=100\), \(K=200\), and exact enumeration,
respectively.

\section{Conclusion}

We introduced PWAL, a pairwise logical method that averages logical
resistance over alternative cross-formula atom-link configurations, rather than committing to a single highest-confidence
configuration.

Across five tasks spanning missing-premise, missing-claim, and abductive
selection, PWAL achieves higher mean strict accuracy and lower mean tie rates
than \textsc{Top-Link}. PWAL also achieves higher mean EAcc on all five tasks.
\textsc{Direct NLI} remains the stronger predictive reference, whereas PWAL
exposes the formulae, link configurations, and resistance components
underlying each decision.

The framework is limited to two candidates, one fixed AMR-derived
propositional representation per statement, and independent uniform
distributions over local link choices. It also inherits errors from AMR
parsing, AMR-to-logic compilation, and NLI-based atom-link construction, while
its Monte Carlo estimates vary with the sampling budget and seed. Future work will study
alternative structured representations and joint uncertainty over
representations and links, dependent or learned link distributions, more
efficient exact inference and adaptive sampling, absolute verification of
individual candidates, and selection among more than two candidates.
\bibliography{aaai2027}

\clearpage
\setcounter{secnumdepth}{2}
\makeatletter
\@addtoreset{definition}{section}
\@addtoreset{example}{section}
\makeatother
\renewcommand{\thedefinition}{\thesection.\arabic{definition}}
\renewcommand{\theexample}{\thesection.\arabic{example}}
\numberwithin{table}{section}
\numberwithin{figure}{section}
\appendix
\section{Dataset Construction and Validity}
\label{sec:supp_dataset_construction}

Every evaluation instance contains two candidate-specific source--target
pairs, \((S_A,T_A)\) and \((S_B,T_B)\), and a gold candidate
\(y\in\{A,B\}\). Table~\ref{tab:dataset_mappings} summarizes the mapping from
each released dataset to this interface, the candidate-construction rule, the
development and test sizes, and the cluster unit. Topic, debate, prompt,
stance, and essay identifiers are used only to construct examples, prevent
split leakage, and define bootstrap clusters. They are not included in
\(S_c\), \(T_c\), \textsc{Direct NLI} input, or symbolic solver input.

\subsection{Pair Mappings and Dataset Splits}

We retain each complete CDED evidence passage as one source text unit because
CDED supervision is passage-level; sentence-level segmentation would
introduce an additional unannotated aggregation choice.

For \(\alpha\)NLI, NumPy seed 1129 selects 400 examples from the official
development pool and 400 from the official test pool after exact duplicate
removal. ARCT development combines 205 deduplicated official-development rows
with 195 deduplicated official-training rows under the same seed; test retains
all 444 official-test rows in official order. Both tasks retain the released
candidate order and have no complete-example or candidate-pair overlap between
development and test.

The three derived tasks use deterministic distractor rules. CDED first prefers
the same released evidence type and then minimizes token-length difference,
subject to token Jaccard similarity below \(0.90\); its token units are
lower-cased ASCII alphanumeric substrings. For iDebate, seed 2026 selects 400
eligible test claims and assigns same-debate distractors injectively by
minimizing total and then maximum token-length difference; its token units are
case-folded Unicode word tokens with optional internal apostrophes. AAE2
minimizes claim-length difference subject to nonidentity, token Jaccard below
\(0.90\), no graph reachability, and no support-evidence overlap; indirect
ancestors and attack neighbors are excluded from its source. Its tokenizer
additionally applies Unicode NFKC normalization before case folding. Residual
distractor-assignment ties are resolved deterministically; candidate order is
fixed so that gold positions are balanced 1:1 for all three derived tasks.

CDED development and test use 19 and 39 disjoint topics, respectively.
iDebate development combines 166 eligible official-development claims with
234 official-training claims; test contains 400 of the 417 eligible
non-singleton official-test claims. AAE2 development contains 200 examples
from 78 official-training essays; its 350-example test set uses 127 disjoint
essays and comprises 103 official-test and 247 official-training targets.
Topic, debate, and essay identifiers are disjoint across the corresponding
development and test splits, and the task-specific overlap checks find no
cross-split leakage.

\paragraph{iDebate qualification.}
The same-debate distractor is topic-controlled but lacks an iDebate annotation
establishing that it is unsupported by the source statement collection. It
may therefore receive partial support. This ambiguity is not used for
filtering or scoring, and debate-clustered intervals account for repeated
examples from one debate.

\begin{table*}[!t]
\centering
\begin{tabular}{@{}
>{\raggedright\arraybackslash}p{0.08\textwidth}
>{\raggedright\arraybackslash}p{0.29\textwidth}
>{\raggedright\arraybackslash}p{0.34\textwidth}
>{\raggedright\arraybackslash}p{0.19\textwidth}
@{}}
\toprule
Dataset & Candidate pair \((S_c,T_c)\) & Candidate construction &
Dev/Test; cluster \\
\midrule
\(\alpha\)NLI
\cite{DBLP:journals/corr/abs-1908-05739}
&
\(S_c=(O_1,H_c)\), \(T_c=(O_2)\)
&
The two released hypotheses are retained in their released order.
&
400/400; example \\
\midrule
ARCT
\cite{Habernal.et.al.2018.NAACL.ARCT}
&
\(S_c=(R,W_c)\), \(T_c=(C)\)
&
The two released warrants are retained.
&
400/444; example \\
\midrule
CDED
\cite{rinott-etal-2015-show}
&
\(S_c=(H_c)\), \(T_c=(C)\)
&
Same-topic evidence annotated as evidence for another claim but not
annotated as evidence for \(C\).
&
200/400; topic \\
\midrule
iDebate
\cite{wang-ling-2016-neural}
&
\(S_c=(P_1,\ldots,P_m)\), \(T_c=(C_c)\)
&
A different central claim from the same debate.
&
400/400; debate \\
\midrule
AAE2
\cite{stab-gurevych-2017-parsing}
&
\(S_c=(P_1,\ldots,P_m)\), \(T_c=(C_c)\)
&
An opposite-stance Claim from the same essay with a graph-disjoint support
subtree.
&
200/350; essay \\
\bottomrule
\end{tabular}
\caption{Source--target mappings, split sizes, and cluster units.
Dataset subsampling is without replacement. Candidate positions are
balanced for the three derived tasks (CDED, iDebate, and AAE2);
\(\alpha\)NLI and ARCT
retain their released candidate order. Parentheses denote ordered statement
collections, including singleton targets.}
\label{tab:dataset_mappings}
\end{table*}

\subsection{AAE2 Graph Extraction}

\begin{figure*}[!t]
\centering
\includegraphics[width=\textwidth]{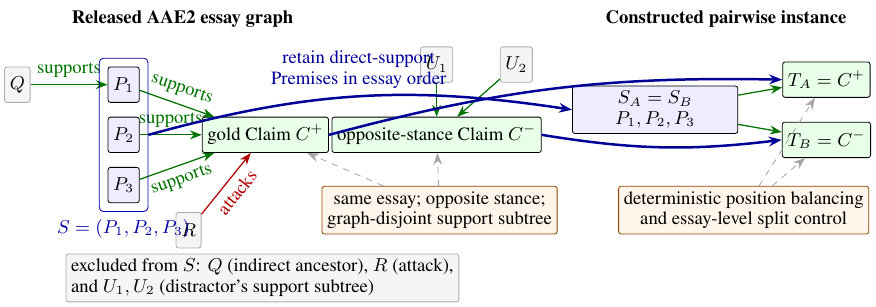}
\caption{AAE2 extraction. The source contains every Premise with a direct
support edge to the gold Claim and no other argument component. The
opposite-stance distractor is selected from the same essay but has a
graph-disjoint support subtree.}
\label{fig:aae2_construction}
\end{figure*}

AAE2 requires an explicit distinction between the released essay graph and
the text passed to the decoder. For a gold Claim \(C^+\), we retain only
Premises whose outgoing relation is a direct support edge to \(C^+\).
Premises that support one of those Premises, components that attack \(C^+\),
and the support subtree of the opposite-stance Claim \(C^-\) remain outside
the source. Because stance is defined relative to the essay's major claim,
\(C^-\) is not assumed to be the logical negation of \(C^+\).
Figure~\ref{fig:aae2_construction} shows this extraction.

The resulting AAE2 sources contain 2.59 Premises on average in development
and 2.57 in test; both medians are 2 and both maxima are 7. Development and
test use disjoint essays. To comply with the source archive's redistribution
terms, we do not redistribute the full derived text.

\subsection{Automatic Construction Audit}

For the automatic checks reported in
Table~\ref{tab:supp_derived_task_audit}, each string is normalized with
Unicode NFKC and case folding; Unicode alphanumeric runs excluding underscores
are then joined with single spaces, so punctuation is discarded.

\begin{table}[t]
\centering
\begin{tabular}{@{}lrrrrrr@{}}
\toprule
Dataset & \(N\) & Clust. & Split ov. & Dup. & Ident. & Verb. \\
\midrule
CDED & 400 & 39 & 0 & 0 & 0 & 0 \\
iDebate & 400 & 128 & 0 & 0 & 0 & 1 \\
AAE2 & 350 & 127 & 0 & 0 & 0 & 0 \\
\bottomrule
\end{tabular}
\caption{Automatic audit of the three derived pairwise tasks. ``Clust.'' is
the number of test clusters; ``Split ov.'' counts development--test cluster
overlap; ``Dup.'' counts duplicate complete examples; ``Ident.'' counts
within-example identical candidates; and ``Verb.'' counts examples in which
the normalized gold-candidate text occurs as a substring of the normalized
source. A complete example includes its source--target context and both
ordered candidates.}
\label{tab:supp_derived_task_audit}
\end{table}

The audit finds no duplicate complete examples, identical within-example
candidates, or development--test cluster overlap. For AAE2,
revalidation of all 550 development and test examples against the released
relation graphs additionally finds no path from a selected source Premise to
its distractor and no overlap with the distractor's support subtree.

One iDebate test example contains the normalized gold-claim string as a
substring of the normalized source. We retain the test set rather than
replacing the item after
observing results. Excluding it changes each reported test accuracy, including
PWAL's ten-seed mean accuracy, by at most \(0.17\) percentage points and does
not alter any comparison.

\subsection{AI-Assisted Validity Audit with Author Verification}
\label{sec:supp_manual_validity_audit}

We additionally audited a fixed sample of 50 CDED, 50 iDebate, and 100 AAE2
test examples. Within each dataset, seed 2026 fixes stratum quotas proportional
to their frequencies in the fixed test set and a deterministic within-stratum
order, with distinct topic, debate, or essay clusters preferred before repeats.
CDED strata combine evidence type and gold position; iDebate strata use gold
position; and AAE2 strata combine stance, gold position, and whether the
distractor has an incoming support subtree. OpenAI GPT-5.6-SOL with maximum
reasoning effort~\cite{openai2026gpt56} produced a first-pass semantic
assessment from the displayed
source or context and the two candidates. One author then reviewed all 200
items and confirmed the first-pass labels. This is an author
verification of an AI-assisted audit, not an independent multi-annotator
study, so we do not report inter-annotator agreement. The first pass was
strictly blinded for 197 items; the gold/distractor role was inadvertently
exposed for one item per dataset. Restricting the summary to the 197 strictly
blinded items gives gold-or-both counts of \(47/49\), \(49/49\), and
\(93/99\), respectively. Table~\ref{tab:supp_manual_validity_audit} reports
the author-confirmed outcomes for the complete 200-item sample.

\paragraph{Audit instruction.}
The complete task-aware instruction used for the first pass was:
\begin{quote}
\raggedright
You are conducting a validity audit of derived pairwise examples. For each
item, use only the displayed task context or source and Candidates A and B.
Do not consult gold labels, adjudication keys, model predictions, or solver
outputs, and do not replace examples after review.

For CDED, determine whether each candidate provides evidence supporting the
displayed claim. For iDebate and AAE2, determine whether the displayed source
supports each candidate claim.

Assign \texttt{supported\_candidate} as exactly one of:\\
\texttt{A} --- only Candidate A is supported;\\
\texttt{B} --- only Candidate B is supported;\\
\texttt{both} --- both are plausibly supported;\\
\texttt{neither} --- neither is supported;\\
\texttt{unclear} --- the displayed text does not permit a clear judgment.

Also assign confidence as \texttt{high}, \texttt{medium}, or \texttt{low},
and give one concise sentence explaining the judgment.

For AAE2 only, assign \texttt{opposite\_stance\_contrast} as exactly one of:\\
\texttt{meaningful\_contrast};\\
\texttt{topic\_related\_not\_contrast};\\
\texttt{unrelated};\\
\texttt{unclear}.
\end{quote}

\begin{table*}[!t]
\centering
\begin{tabular}{lrrrrrr}
\toprule
Dataset & \(N\) & Only gold & Both & Only distractor & Neither & Gold or both \\
\midrule
CDED    & 50  & 39 & 9  & 0 & 2 & 48 (96\%) \\
iDebate & 50  & 37 & 13 & 0 & 0 & 50 (100\%) \\
AAE2    & 100 & 91 & 3  & 4 & 2 & 94 (94\%) \\
\bottomrule
\end{tabular}
\caption{Author-confirmed outcomes of the AI-assisted validity audit.
``Gold or both'' counts examples for which the gold candidate alone or both
candidates were plausible. No item was unclear.}
\label{tab:supp_manual_validity_audit}
\end{table*}

For the AAE2 contrast check, 92 of 100 distractors were judged meaningful
opposite-stance contrasts; the other 8 were topic-related but did not form a
meaningful contrast. These audit labels were not used to modify the test sets
or select any method setting.

\FloatBarrier
\section{Evaluation Protocol and Development Decisions}
\label{sec:supp_protocol}

\subsection{Evaluation Settings and Compute Environment}

Experiments were run on 64-bit Windows with an Intel Core i5-13600KF CPU,
32\,GB RAM, and an NVIDIA GeForce RTX 4090 GPU with 24\,GB memory, using
Python 3.8.18, PyTorch 1.13.1 with CUDA 11.7, Transformers 4.34.0,
Sentence-Transformers 2.2.2, and PySAT 0.1.8.dev9. The common pipeline uses
Structured-BART for AMR parsing~\cite{zhou2021structure,lee2022maximum},
mDeBERTa-v3 for NLI-based atom links
\cite{he2023debertav3,laurer2024mdebertaxnli}, and RC2 in PySAT for Partial
MaxSAT inference~\cite{Ignatiev2018}. \textsc{FH-HardSAT} additionally uses
\texttt{BAAI/bge-small-en-v1.5} for cosine similarity~\cite{bge_embedding}.
For method selection, development labels are used only for
\textsc{FH-HardSAT} threshold selection. Separately, development labels are
used to evaluate the diagnostic resistance-component sweep in
Section~\ref{sec:supp_alpha_sensitivity}. No test label is used to choose the
scoring rule, sampling budget, seed set, ablation, or reference policy.

For an evaluation set of \(N\) examples, let
\(N_{\mathrm{win}},N_{\mathrm{err}},N_{\mathrm{tie}}\) denote correct unique
decisions, errors, and valid score ties, with
\(N=N_{\mathrm{win}}+N_{\mathrm{err}}+N_{\mathrm{tie}}\). The reported
metrics are
\[
\begin{aligned}
\operatorname{Accuracy}&=\frac{N_{\mathrm{win}}}{N},&
\operatorname{TieRate}&=\frac{N_{\mathrm{tie}}}{N},\\
\operatorname{EAcc}&=\frac{N_{\mathrm{win}}+\tfrac12N_{\mathrm{tie}}}{N}.&&
\end{aligned}
\]
Any absolute difference of at most \(\tau_{\mathrm{tie}}=10^{-12}\) between
the two candidate scores is treated as a valid tie. Invalid outputs are a diagnostic
subset of \(N_{\mathrm{err}}\): they receive zero credit and are never
relabeled as ties.

\subsection{\textsc{FH-HardSAT} Threshold Selection}

\textsc{FH-HardSAT} searches its threshold grid on the corresponding
development split and applies the selected pair once to the test split.
Table~\ref{tab:supp_fh_settings} reports the selected thresholds and held-out
test counts without repeating the main-paper metrics.

\begin{table}[t]
\centering
\begin{tabular}{@{}lrrrr@{}}
\toprule
Dataset & \(\tau_c\) & \(\tau_m\) & Dev Acc. (\%) & Test \(W/L/T\) \\
\midrule
$\alpha$NLI & 90 & 0.55 & 11.50 & 44/133/223 \\
ARCT & 100 & 0.60 & 5.25 & 33/26/385 \\
CDED & 100 & 0.60 & 35.50 & 138/63/199 \\
iDebate & 100 & 0.70 & 29.25 & 132/48/220 \\
AAE2 & 100 & 0.60 & 17.50 & 66/67/217 \\
\bottomrule
\end{tabular}
\caption{\textsc{FH-HardSAT} development selection and held-out test counts.
\(\tau_m\) is the BGE cosine-similarity threshold and \(\tau_c\) the NLI
contradiction-confidence threshold (in percent). Test \(W/L/T\) gives correct
unique decisions, errors, and valid ties, respectively; \(L\) includes
unsuccessful evaluations. The corresponding metrics appear in the main
results.}
\label{tab:supp_fh_settings}
\end{table}

\FloatBarrier
\section{Decision Diagnostics and Components}
\label{sec:supp_diagnostics}

\subsection{\textsc{Top-Link}-to-PWAL Transitions}
\label{sec:supp_decision_transitions}

We classify each output as a correct unique decision, an incorrect unique
decision, or a valid tie. \textsc{Top-Link} is deterministic, whereas PWAL
transitions are computed separately for each of the ten \(K=100\) seeds
before aggregation. Table~\ref{tab:supp_toplink_pwal_ties} reports PWAL EAcc
over the \textsc{Top-Link} ties and full-test unique-to-unique repairs and
damages. A tied-subset EAcc of \(50\%\) is the uniform tie-breaking reference,
not a significance threshold.

\begin{table*}[t]
\centering
\begin{tabular}{@{}lrrrr@{}}
\toprule
Dataset & \shortstack{TL ties\\\((N)\)} &
\shortstack{PWAL EAcc\\on TL ties \((\%)\)} & \shortstack{Repairs\\\((N)\)} &
\shortstack{Damages\\\((N)\)} \\
\midrule
\(\alpha\)NLI & 109 & \(49.36\pm1.97\) & \(32.1\pm2.13\) & \(29.4\pm1.84\) \\
ARCT           & 288 & \(51.82\pm1.49\) & \(11.7\pm1.16\) & \(8.7\pm1.77\) \\
CDED           & 54  & \(57.13\pm3.55\) & \(39.0\pm2.40\) & \(18.7\pm0.82\) \\
iDebate        & 20  & \(52.50\pm2.64\) & \(33.9\pm1.97\) & \(32.6\pm1.35\) \\
AAE2           & 26  & \(51.15\pm3.17\) & \(26.6\pm1.07\) & \(23.4\pm1.17\) \\
\bottomrule
\end{tabular}
\caption{\textsc{Top-Link}-to-PWAL transition summary. Tied-subset EAcc is
PWAL EAcc on the deterministic \textsc{Top-Link} ties. Repairs and damages are
full-test mean counts of unique-to-unique transitions from incorrect to
correct and correct to incorrect, respectively. Entries are ten-seed means
\(\pm\) sample standard deviations.}
\label{tab:supp_toplink_pwal_ties}
\end{table*}

CDED has the highest tied-subset EAcc (\(57.13\%\)). Repairs exceed damages on
every dataset, although the net difference is small on \(\alpha\)NLI,
iDebate, and AAE2.

\subsection{Score-Component Error Signatures}
\label{sec:supp_error_decomposition}

For each incorrect unique PWAL decision, we identify which score components
favor the selected candidate. The tension component favors it when the gold
candidate has higher normalized semantic tension; the witness component
favors it when the selected candidate has a higher target-clause witness
ratio. For this diagnostic component attribution only, component differences
within \(10^{-9}\) are treated as zero. This diagnostic tolerance does not
alter the candidate-score tie rule \(\tau_{\mathrm{tie}}=10^{-12}\).

Table~\ref{tab:supp_error_decomposition} reports the resulting
score-component categories for all five datasets.

\begin{table*}[t]
\centering
\begin{tabular}{@{}lrrrr@{}}
\toprule
Dataset & Wrong & Tension & Witness & Both \\
\midrule
\(\alpha\)NLI & \(193.20\pm4.26\) & \(31.73\pm1.72\) & \(15.42\pm2.00\) & \(52.86\pm2.47\) \\
ARCT & \(187.50\pm4.72\) & \(33.07\pm2.17\) & \(47.36\pm2.33\) & \(19.57\pm2.45\) \\
CDED & \(95.70\pm2.75\) & \(16.71\pm2.07\) & \(27.60\pm2.53\) & \(55.69\pm2.90\) \\
iDebate & \(145.20\pm2.62\) & \(8.73\pm1.23\) & \(9.91\pm1.45\) & \(81.35\pm1.54\) \\
AAE2 & \(149.50\pm1.96\) & \(11.24\pm1.31\) & \(8.15\pm1.06\) & \(80.61\pm1.57\) \\
\bottomrule
\end{tabular}
\caption{Score-component attribution of PWAL's incorrect unique decisions at
\(K=100\). Wrong is the mean count of incorrect unique decisions; the
remaining columns are mean percentages. Entries are means \(\pm\) sample
standard deviations over seeds 2026--2035.}
\label{tab:supp_error_decomposition}
\end{table*}

These categories describe score-component signatures, not linguistic causes.
Both components favor the incorrect selection in a majority of errors on
\(\alpha\)NLI, CDED, iDebate, and AAE2; witness-only attribution is the
largest category on ARCT.

\subsection{Resistance-Component Sensitivity}
\label{sec:supp_resistance_score_analysis}
\label{sec:supp_alpha_sensitivity}

To diagnose the relative contribution of the two terms in the fixed
logical-resistance score, we evaluate a coefficient sweep on development data
only. The reported score remains
\[
\rho=\frac{\Delta T}{C_{\max}}-R_{\mathrm{sat}},
\]
where \(\Delta T/C_{\max}\) is normalized semantic tension and
\(R_{\mathrm{sat}}\) is the target-clause witness ratio. The diagnostic sweep
defines
\[
\rho_{\alpha}=\alpha\frac{\Delta T}{C_{\max}}-R_{\mathrm{sat}}
\]
and evaluates every \(\alpha\in\{0,0.05,\ldots,5\}\). This development-only
sweep does not tune the test score: \(\rho_{\alpha}=\rho\) at \(\alpha=1\),
and all reported test results use \(\rho\). All other settings remain fixed:
\textsc{Top-Link} uses its single
highest-confidence link configuration, while PWAL uses \(K=100\) and seeds
2026--2035.

\begin{figure*}[t]
\centering
\includegraphics[width=\textwidth]{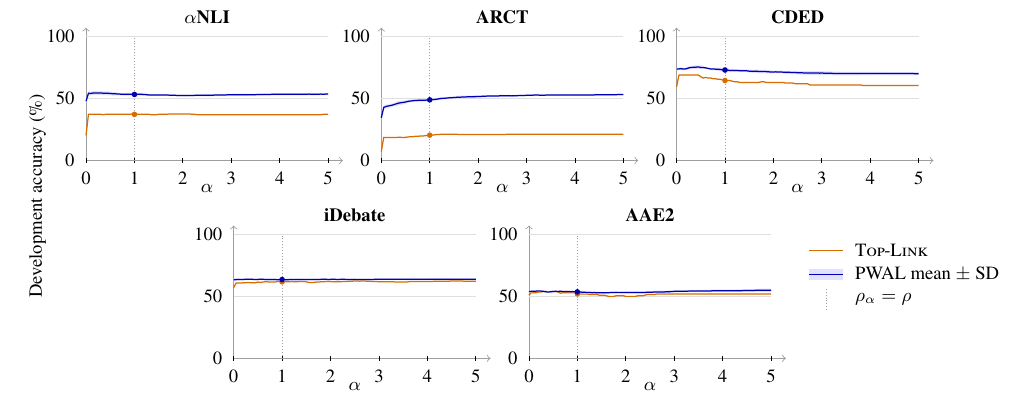}
\caption{Development-only diagnostic of the logical-resistance components
under \(\rho_{\alpha}=\alpha\Delta T/C_{\max}-R_{\mathrm{sat}}\). Curves
contain every point in the fixed grid \(\alpha=0,0.05,\ldots,5\). PWAL curves
are ten-seed means and shaded bands show one sample standard deviation;
\textsc{Top-Link} is deterministic. Filled markers identify the reported score
\(\rho_{\alpha}=\rho\) at \(\alpha=1\).}
\label{fig:supp_resistance_score_analysis}
\end{figure*}

Figure~\ref{fig:supp_resistance_score_analysis} shows that, across the ten
dataset--method curves, the development accuracy obtained with \(\rho\) is
\(0.25\)--\(4.50\) percentage points below the maximum observed on the grid.

\subsection{Component Contrasts}
\label{sec:supp_ablations}

The no-link state permits an unmatched target atom to remain unlinked in a
sampled world. The negative guard allows a negative target literal to count
as witnessed only when its polarity is supported by an exact link whose
source atom is false or by an active contradiction link whose source atom is
true; target inertia alone is insufficient. Each ablation changes only the
named component and keeps
the scoring rule \(\rho\), \(K=100\), seeds 2026--2035, formulae, atom
inventories,
exact links, NLI alternatives, and clause weights fixed. The negative-guard
ablation reuses Full PWAL's sampled worlds for every example--seed pair and
recomputes only witness and resistance terms. Removing no-link changes the
local outcome distribution, so it uses the same seed IDs but resamples under
the modified distribution. The main paper reports the point estimates;
Table~\ref{tab:supp_paired_components} gives paired cluster-aware intervals.

\paragraph{AAE2 stance diagnostic.}
For AAE2, PWAL EAcc is \(56.41\%\) for \texttt{For} claims and \(54.29\%\)
for \texttt{Against} claims, averaged over seeds 2026--2035. This diagnostic
was not used to select any method setting.

\FloatBarrier
\section{Sampling, Exact/\(10^5\)-Capped Reference, and Runtime}
\label{sec:supp_sampling}
\label{sec:supp_exact_reference}

\subsection{Sampling Stability}

For each \(K\in\{10,20,50,100,200\}\), PWAL is evaluated with seeds
2026--2035. Within each seed, the smaller budgets are prefixes of the same
\(K=200\) world stream.
Table~\ref{tab:supp_k_stability} reports the corresponding accuracy, tie-rate,
and EAcc summaries together with the fixed exact/\(10^5\)-capped reference.

\begin{table*}[t]
\centering
\begin{tabular}{@{}lrrrr@{}}
\toprule
Dataset & \(K\) & Accuracy & Tie & EAcc \\
\midrule
\multirow{6}{*}{$\alpha$NLI} & 10 & 47.83 $\pm$ 1.62 & 5.63 $\pm$ 0.48 & 50.64 $\pm$ 1.68 \\
 & 20 & 49.08 $\pm$ 1.83 & 4.28 $\pm$ 0.59 & 51.21 $\pm$ 1.80 \\
 & 50 & 49.30 $\pm$ 0.86 & 3.18 $\pm$ 0.51 & 50.89 $\pm$ 0.76 \\
 & 100 & 49.55 $\pm$ 0.96 & 2.15 $\pm$ 0.50 & 50.63 $\pm$ 0.98 \\
 & 200 & 49.92 $\pm$ 1.16 & 1.48 $\pm$ 0.25 & 50.66 $\pm$ 1.17 \\
 & Ref. & 49.75 & 3.25 & 51.38 \\
\midrule
\multirow{6}{*}{ARCT} & 10 & 47.52 $\pm$ 1.42 & 12.21 $\pm$ 0.91 & 53.63 $\pm$ 1.39 \\
 & 20 & 48.51 $\pm$ 1.86 & 9.19 $\pm$ 0.52 & 53.11 $\pm$ 1.79 \\
 & 50 & 49.66 $\pm$ 2.27 & 7.59 $\pm$ 0.51 & 53.46 $\pm$ 2.15 \\
 & 100 & 50.90 $\pm$ 1.12 & 6.87 $\pm$ 0.41 & 54.34 $\pm$ 1.07 \\
 & 200 & 52.00 $\pm$ 1.59 & 6.71 $\pm$ 0.51 & 55.36 $\pm$ 1.47 \\
 & Ref. & 48.20 & 14.41 & 55.41 \\
\midrule
\multirow{6}{*}{CDED} & 10 & 71.53 $\pm$ 1.23 & 4.73 $\pm$ 0.49 & 73.89 $\pm$ 1.17 \\
 & 20 & 71.93 $\pm$ 1.23 & 4.18 $\pm$ 0.54 & 74.01 $\pm$ 1.31 \\
 & 50 & 72.68 $\pm$ 1.03 & 3.33 $\pm$ 0.37 & 74.34 $\pm$ 0.95 \\
 & 100 & 73.50 $\pm$ 0.79 & 2.58 $\pm$ 0.29 & 74.79 $\pm$ 0.73 \\
 & 200 & 73.55 $\pm$ 0.45 & 2.28 $\pm$ 0.30 & 74.69 $\pm$ 0.47 \\
 & Ref. & 74.25 & 1.50 & 75.00 \\
\midrule
\multirow{6}{*}{iDebate} & 10 & 63.23 $\pm$ 0.58 & 0.15 $\pm$ 0.13 & 63.30 $\pm$ 0.64 \\
 & 20 & 63.73 $\pm$ 0.99 & 0.08 $\pm$ 0.12 & 63.76 $\pm$ 0.99 \\
 & 50 & 63.65 $\pm$ 1.06 & 0.08 $\pm$ 0.12 & 63.69 $\pm$ 1.04 \\
 & 100 & 63.70 $\pm$ 0.65 & 0.00 $\pm$ 0.00 & 63.70 $\pm$ 0.65 \\
 & 200 & 63.50 $\pm$ 0.31 & 0.00 $\pm$ 0.00 & 63.50 $\pm$ 0.31 \\
 & Ref. & 63.25 & 0.00 & 63.25 \\
\midrule
\multirow{6}{*}{AAE2} & 10 & 53.94 $\pm$ 0.81 & 3.09 $\pm$ 0.23 & 55.49 $\pm$ 0.83 \\
 & 20 & 54.14 $\pm$ 0.93 & 2.97 $\pm$ 0.20 & 55.63 $\pm$ 0.92 \\
 & 50 & 54.51 $\pm$ 0.92 & 2.86 $\pm$ 0.00 & 55.94 $\pm$ 0.92 \\
 & 100 & 54.43 $\pm$ 0.56 & 2.86 $\pm$ 0.00 & 55.86 $\pm$ 0.56 \\
 & 200 & 54.43 $\pm$ 0.49 & 2.89 $\pm$ 0.09 & 55.87 $\pm$ 0.49 \\
 & Ref. & 54.29 & 2.86 & 55.71 \\
\bottomrule
\end{tabular}
\caption{PWAL stability and exact/\(10^5\)-capped reference results.
Finite-\(K\) cells are ten-seed mean percentages \(\pm\) sample standard
deviations in percentage points; Ref. is a fixed percentage.}
\label{tab:supp_k_stability}
\end{table*}

At \(K=100\), the across-seed accuracy standard deviation is at most \(1.12\)
percentage points. Increasing \(K\) from 100 to 200 changes mean accuracy and
mean tie rate by at most \(1.10\) and \(0.68\) percentage points, respectively.

\subsection{Reference Coverage and Approximation}

For candidate \(c\) with world set \(\mathcal W_c\), let
\(N_c^{\mathrm{world}}=|\mathcal W_c|\). The reference enumerates all worlds
when \(N_c^{\mathrm{world}}\leq10^5\) and otherwise draws \(10^5\) worlds
uniformly without replacement using seed 2026. A pair is fully exact only when
both candidates are enumerated. Table~\ref{tab:supp_reference_summary} reports the
world-space quantities that determine exact coverage; the corresponding
reference outcomes are included in Table~\ref{tab:supp_k_stability}.

\begin{table*}[t]
\centering
\begin{tabular}{@{}lrrrr@{}}
\toprule
Dataset & Exact pairs & Capped pairs & Median & P95 \\
\midrule
\(\alpha\)NLI & 357 & 43  & 294     & 205,920 \\
ARCT           & 443 & 1   & 42      & 1,848 \\
CDED           & 216 & 184 & 6,210   & \(7.47\times10^{10}\) \\
iDebate        & 146 & 254 & 37,440  & \(5.00\times10^{11}\) \\
AAE2           & 93  & 257 & 170,586 & \(5.40\times10^{13}\) \\
\bottomrule
\end{tabular}
\caption{Exact/\(10^5\)-capped reference coverage and world-space scale.
World-count summaries are over candidate sides before capping; reference
outcomes are included in Table~\ref{tab:supp_k_stability} and summarized in
the main paper. A capped pair has at least one candidate with
\(N_c^{\mathrm{world}}>10^5\); P95 denotes the 95th percentile.}
\label{tab:supp_reference_summary}
\end{table*}

For a dataset of \(N\) examples, let
\(\mathcal S=\{2026,\ldots,2035\}\) and let
\[
\overline\mu^{(100)}_{i,c}
=\frac{1}{|\mathcal S|}\sum_{s\in\mathcal S}
\widehat\mu^{(100,s)}_{i,c}
\]
be the ten-seed mean \(K=100\) score for example \(i\) and candidate
\(c\in\{A,B\}\); let \(\mu^{\mathrm{ref}}_{i,c}\) be its reference score.
Define the corresponding candidate-score margins as
\[
m_i^{(100)}=\overline\mu^{(100)}_{i,B}-\overline\mu^{(100)}_{i,A},
\qquad
m_i^{\mathrm{ref}}=\mu^{\mathrm{ref}}_{i,B}-\mu^{\mathrm{ref}}_{i,A}.
\]
The candidate-score and margin mean absolute errors (MAEs) are
\[
\begin{aligned}
\operatorname{MAE}_{\mathrm{score}}
&=\frac{1}{2N}\sum_{i=1}^{N}\sum_{c\in\{A,B\}}
\left|\overline\mu^{(100)}_{i,c}-\mu^{\mathrm{ref}}_{i,c}\right|,\\
\operatorname{MAE}_{\mathrm{margin}}
&=\frac{1}{N}\sum_{i=1}^{N}
\left|m_i^{(100)}-m_i^{\mathrm{ref}}\right|.
\end{aligned}
\]
Decision agreement is the percentage of examples for which the mean-score and
reference evaluations return the same \(A/B/\texttt{Tie}\) decision.
Tie-status agreement is the percentage for which both evaluations are either
tied or unique, irrespective of which candidate wins when both are unique.
Table~\ref{tab:supp_exact_diagnostics} reports these two agreements together
with the two errors and the per-seed attribution of sampled incorrect unique
decisions.

\begin{table*}[t]
\centering
\begin{tabular}{@{}lrrrr@{}}
\toprule
\multicolumn{5}{c}{\textbf{(a) Ten-seed mean-score approximation}} \\
\midrule
Dataset & Score MAE & Margin MAE & Decision agr. & Tie-status agr. \\
\midrule
\(\alpha\)NLI & 0.0080 & 0.0117 & 95.25 & 97.25 \\
ARCT & 0.0078 & 0.0111 & 88.51 & 91.67 \\
CDED & 0.0069 & 0.0105 & 98.75 & 99.25 \\
iDebate & 0.0089 & 0.0122 & 99.00 & 100.00 \\
AAE2 & 0.0066 & 0.0094 & 99.71 & 100.00 \\
\midrule
\multicolumn{5}{c}{\textbf{(b) Per-seed exact attribution}} \\
\midrule
Dataset & \shortstack{Sample--exact\\decision agr.} &
\shortstack{Wrong\(\to\)exact\\correct} &
\shortstack{Wrong\(\to\)exact\\tie} &
\shortstack{Wrong\(\to\)exact\\wrong} \\
\midrule
\(\alpha\)NLI & \(89.47\pm1.55\) & \(6.95\pm2.11\) & \(3.04\pm0.61\) & \(90.02\pm2.34\) \\
ARCT & \(83.02\pm1.34\) & \(10.95\pm1.49\) & \(10.33\pm1.77\) & \(78.72\pm2.35\) \\
CDED & \(96.34\pm0.99\) & \(5.36\pm2.43\) & \(0.56\pm0.90\) & \(94.08\pm2.76\) \\
iDebate & \(97.67\pm1.22\) & \(2.78\pm1.98\) & \(0.00\pm0.00\) & \(97.22\pm1.98\) \\
AAE2 & \(98.39\pm1.54\) & \(1.10\pm1.43\) & \(0.00\pm0.00\) & \(98.90\pm1.43\) \\
\bottomrule
\end{tabular}
\caption{Reference diagnostics. The first block compares decisions formed
from ten-seed \(K=100\) mean scores with the exact/\(10^5\)-capped reference.
In panel (a), MAEs are in resistance-score units and the agreement columns are
percentages. The second block compares every \(K=100\) seed separately
with exact marginalization on fully exact pairs; entries are mean
percentages \(\pm\) sample standard deviations over seeds 2026--2035.
Its final three columns partition sampled incorrect unique decisions into
those that become correct, become a tie, or remain wrong.}
\label{tab:supp_exact_diagnostics}
\end{table*}

Candidate-score MAE is \(0.0066\)--\(0.0089\), margin MAE is
\(0.0094\)--\(0.0122\), and mean-score decision agreement is
\(88.51\%\)--\(99.71\%\).

Within the fully exact subsets, \(78.72\%\)--\(98.90\%\) of sampled incorrect
unique decisions remain wrong under exact marginalization;
\(1.10\%\)--\(10.95\%\) become correct, and the remainder become exact ties.
For capped candidates, the reference remains a deterministic uniform
\(10^5\)-world estimate rather than an exact expectation.

\subsection{Runtime}

Runtime isolates logical scoring on the same fixed set of 20 fully enumerable
test pairs per dataset for every configuration. The pairs are selected once by
a fixed label-independent ranking of sample IDs. Each per-example value is the
median of three timed runs after one warm-up; finite-\(K\) configurations use
seed 2026. Timing begins from prepared logical states and excludes parsing,
translation, NLI, and cache construction.

\begin{table*}[t]
\centering
\begin{tabular}{@{}llrrr@{}}
\toprule
Dataset & Configuration & Mean & Median & P95 \\
\midrule
\multirow{7}{*}{$\alpha$NLI}
& \textsc{Top-Link} & 1.66 & 1.67 & 2.16 \\
& PWAL \(K=10\) & 14.58 & 14.37 & 21.71 \\
& PWAL \(K=20\) & 27.31 & 26.19 & 41.30 \\
& PWAL \(K=50\) & 68.19 & 66.75 & 101.80 \\
& PWAL \(K=100\) & 135.59 & 133.95 & 203.65 \\
& PWAL \(K=200\) & 280.89 & 283.08 & 411.32 \\
& Exact enumeration & 7,608.02 & 354.35 & 33,502.77 \\
\midrule
\multirow{7}{*}{ARCT}
& \textsc{Top-Link} & 1.41 & 1.55 & 1.84 \\
& PWAL \(K=10\) & 12.08 & 11.88 & 17.47 \\
& PWAL \(K=20\) & 25.92 & 24.83 & 34.25 \\
& PWAL \(K=50\) & 61.86 & 63.11 & 85.55 \\
& PWAL \(K=100\) & 120.22 & 119.82 & 170.60 \\
& PWAL \(K=200\) & 250.90 & 260.40 & 333.48 \\
& Exact enumeration & 526.31 & 53.88 & 1,694.38 \\
\midrule
\multirow{7}{*}{CDED}
& \textsc{Top-Link} & 1.68 & 1.66 & 2.28 \\
& PWAL \(K=10\) & 14.63 & 15.35 & 20.15 \\
& PWAL \(K=20\) & 25.43 & 25.16 & 36.28 \\
& PWAL \(K=50\) & 62.11 & 61.03 & 89.96 \\
& PWAL \(K=100\) & 126.53 & 126.21 & 175.92 \\
& PWAL \(K=200\) & 262.79 & 267.03 & 379.23 \\
& Exact enumeration & 12,873.73 & 1,792.32 & 61,171.62 \\
\midrule
\multirow{7}{*}{iDebate}
& \textsc{Top-Link} & 1.57 & 1.48 & 2.22 \\
& PWAL \(K=10\) & 15.38 & 13.99 & 23.13 \\
& PWAL \(K=20\) & 30.79 & 28.66 & 46.39 \\
& PWAL \(K=50\) & 72.23 & 67.55 & 101.76 \\
& PWAL \(K=100\) & 142.47 & 141.32 & 203.66 \\
& PWAL \(K=200\) & 289.34 & 278.08 & 386.68 \\
& Exact enumeration & 14,764.91 & 5,567.09 & 78,545.72 \\
\midrule
\multirow{7}{*}{AAE2}
& \textsc{Top-Link} & 1.10 & 1.06 & 1.56 \\
& PWAL \(K=10\) & 9.51 & 8.53 & 13.45 \\
& PWAL \(K=20\) & 19.60 & 17.38 & 33.37 \\
& PWAL \(K=50\) & 48.49 & 42.07 & 81.37 \\
& PWAL \(K=100\) & 94.04 & 85.63 & 145.66 \\
& PWAL \(K=200\) & 188.33 & 165.96 & 283.57 \\
& Exact enumeration & 7,886.06 & 1,823.10 & 24,037.05 \\
\bottomrule
\end{tabular}
\caption{Controlled logical-scoring runtime on 20 fully enumerable pairs per
dataset (ms/example). P95 denotes the 95th percentile.}
\label{tab:supp_runtime}
\end{table*}

Table~\ref{tab:supp_runtime} shows that mean \(K=100\) runtime is
\(94.04\)--\(142.47\) ms per example, compared with \(1.10\)--\(1.68\) ms for
\textsc{Top-Link}. \(K=200\) approximately doubles the \(K=100\) runtime,
while exact enumeration requires \(526.31\)--\(14{,}764.91\) ms on the same
pairs.

\FloatBarrier
\section{Paired and Clustered Statistical Inference}
\label{sec:supp_statistics}

\begin{table*}[t]
\centering
\begin{tabular}{@{}llccc@{}}
\toprule
\multicolumn{5}{c}{\textbf{(a) Main-method contrasts}} \\
\midrule
Dataset & Contrast & \(\Delta\)Acc [95\% CI] & \(\Delta\)EAcc [95\% CI] & Tie reduction [95\% CI] \\
\midrule
$\alpha$NLI & PWAL \(-\) \textsc{Direct NLI} & -8.20 [-14.52, -1.60]\(^{\dagger}\) & -7.12 [-13.46, -0.57]\(^{\dagger}\) & -2.15 [-3.38, -1.10]\(^{\dagger}\) \\
$\alpha$NLI & PWAL \(-\) \textsc{Top-Link} & +13.05 [+8.45, +17.68]\(^{\dagger}\) & +0.50 [-3.70, +4.69] & +25.10 [+21.00, +29.38]\(^{\dagger}\) \\
$\alpha$NLI & PWAL \(-\) \textsc{FH-HardSAT} & +38.55 [+33.27, +43.85]\(^{\dagger}\) & +11.75 [+6.41, +17.14]\(^{\dagger}\) & +53.60 [+48.62, +58.67]\(^{\dagger}\) \\
\midrule
ARCT & PWAL \(-\) \textsc{Direct NLI} & -19.14 [-25.18, -13.11]\(^{\dagger}\) & -15.71 [-21.69, -9.90]\(^{\dagger}\) & -6.87 [-9.14, -4.73]\(^{\dagger}\) \\
ARCT & PWAL \(-\) \textsc{Top-Link} & +30.86 [+26.80, +34.91]\(^{\dagger}\) & +1.86 [-1.58, +5.24] & +58.00 [+53.33, +62.50]\(^{\dagger}\) \\
ARCT & PWAL \(-\) \textsc{FH-HardSAT} & +43.47 [+38.81, +48.18]\(^{\dagger}\) & +3.55 [-0.72, +7.92] & +79.84 [+75.90, +83.58]\(^{\dagger}\) \\
\midrule
CDED & PWAL \(-\) \textsc{Direct NLI} & -6.50 [-11.35, -1.78]\(^{\dagger}\) & -5.21 [-9.99, -0.51]\(^{\dagger}\) & -2.57 [-4.12, -1.21]\(^{\dagger}\) \\
CDED & PWAL \(-\) \textsc{Top-Link} & +11.50 [+7.70, +15.45]\(^{\dagger}\) & +6.04 [+2.28, +9.94]\(^{\dagger}\) & +10.93 [+7.45, +14.67]\(^{\dagger}\) \\
CDED & PWAL \(-\) \textsc{FH-HardSAT} & +39.00 [+32.45, +45.01]\(^{\dagger}\) & +15.41 [+9.99, +20.58]\(^{\dagger}\) & +47.17 [+42.48, +51.99]\(^{\dagger}\) \\
\midrule
iDebate & PWAL \(-\) \textsc{Direct NLI} & -8.80 [-13.76, -3.95]\(^{\dagger}\) & -8.80 [-13.76, -3.95]\(^{\dagger}\) & +0.00 [+0.00, +0.00] \\
iDebate & PWAL \(-\) \textsc{Top-Link} & +2.95 [-1.15, +6.94] & +0.45 [-3.55, +4.32] & +5.00 [+2.77, +7.48]\(^{\dagger}\) \\
iDebate & PWAL \(-\) \textsc{FH-HardSAT} & +30.70 [+25.64, +35.66]\(^{\dagger}\) & +3.20 [-1.17, +7.40] & +55.00 [+49.88, +59.90]\(^{\dagger}\) \\
\midrule
AAE2 & PWAL \(-\) \textsc{Direct NLI} & -16.14 [-22.96, -9.40]\(^{\dagger}\) & -14.71 [-21.36, -8.16]\(^{\dagger}\) & -2.86 [-4.58, -1.22]\(^{\dagger}\) \\
AAE2 & PWAL \(-\) \textsc{Top-Link} & +3.29 [-0.38, +6.99] & +1.00 [-2.54, +4.54] & +4.57 [+2.51, +6.92]\(^{\dagger}\) \\
AAE2 & PWAL \(-\) \textsc{FH-HardSAT} & +35.57 [+29.70, +41.43]\(^{\dagger}\) & +6.00 [+0.64, +11.22]\(^{\dagger}\) & +59.14 [+53.42, +64.86]\(^{\dagger}\) \\
\midrule
\multicolumn{5}{c}{\textbf{(b) Component contrasts}} \\
\midrule
Dataset & Contrast & \(\Delta\)Acc [95\% CI] & \(\Delta\)EAcc [95\% CI] & Tie reduction [95\% CI] \\
\midrule
$\alpha$NLI & Full \(-\) w/o no-link & +3.40 [+0.85, +6.02]\(^{\dagger}\) & -1.51 [-3.91, +0.85] & +9.83 [+7.07, +12.70]\(^{\dagger}\) \\
$\alpha$NLI & Full \(-\) w/o negative guard & +0.07 [-1.03, +1.20] & -1.16 [-2.26, -0.15]\(^{\dagger}\) & +2.48 [+1.15, +4.08]\(^{\dagger}\) \\
\midrule
ARCT & Full \(-\) w/o no-link & +3.13 [+0.99, +5.36]\(^{\dagger}\) & -2.09 [-3.95, -0.24]\(^{\dagger}\) & +10.45 [+7.79, +13.33]\(^{\dagger}\) \\
ARCT & Full \(-\) w/o negative guard & +10.32 [+6.98, +13.67]\(^{\dagger}\) & -2.20 [-4.97, +0.54] & +25.02 [+21.15, +29.05]\(^{\dagger}\) \\
\midrule
CDED & Full \(-\) w/o no-link & +2.00 [+0.08, +4.00]\(^{\dagger}\) & +0.50 [-1.15, +2.14] & +3.00 [+1.46, +4.84]\(^{\dagger}\) \\
CDED & Full \(-\) w/o negative guard & +3.90 [+1.73, +6.17]\(^{\dagger}\) & +1.07 [-0.23, +2.25] & +5.65 [+2.98, +8.64]\(^{\dagger}\) \\
\midrule
iDebate & Full \(-\) w/o no-link & +0.22 [-0.75, +1.21] & +0.21 [-0.77, +1.19] & +0.03 [+0.00, +0.15] \\
iDebate & Full \(-\) w/o negative guard & +0.07 [-0.78, +0.95] & -0.09 [-1.01, +0.81] & +0.33 [+0.00, +0.97] \\
\midrule
AAE2 & Full \(-\) w/o no-link & +0.54 [-1.26, +2.38] & +0.10 [-1.59, +1.79] & +0.89 [+0.00, +1.97] \\
AAE2 & Full \(-\) w/o negative guard & +0.26 [-1.37, +2.08] & +0.50 [-0.91, +2.05] & -0.49 [-2.17, +1.18] \\
\bottomrule
\end{tabular}
\caption{Paired two-way bootstrap contrasts in percentage points. Brackets
are pointwise two-sided 95\% bootstrap percentile intervals; \(\dagger\) marks an
interval that excludes zero.}
\label{tab:supp_paired_main}
\label{tab:supp_paired_components}
\end{table*}

The contrasts in Table~\ref{tab:supp_paired_main} use the following bootstrap
protocol. Each reported point estimate averages per-example utility over the
ten complete \(K=100\) seed runs. Each of 10,000 bootstrap replicates resamples
complete task-specific clusters and whole seed runs with replacement while
preserving method pairing. Cluster units are example for \(\alpha\)NLI and
ARCT, topic for CDED, debate for iDebate, and essay for AAE2. Reported
intervals are pointwise two-sided 95\% bootstrap percentile intervals;
the ten seeds are not treated as ten independent test sets, and an interval
containing zero is treated as statistically inconclusive.

Strict-accuracy utility is one only for a correct unique decision; EAcc
utility is one for a correct unique decision and one half for a valid score
tie; tie utility is one only for a valid score tie. For accuracy and EAcc, a
positive main-method contrast favors PWAL and a positive component contrast
favors Full. Tie reduction reverses the subtraction order, so a positive
value means fewer ties for PWAL or Full.

PWAL's strict-accuracy interval relative to \textsc{Top-Link} excludes zero
on \(\alpha\)NLI, ARCT, and CDED but includes zero on iDebate and AAE2. For
component contrasts, the no-link accuracy interval excludes zero on
\(\alpha\)NLI, ARCT, and CDED, while the negative-guard interval excludes
zero on ARCT and CDED.
\FloatBarrier
\section{Translator Specification}
\label{sec:supp_implementation}

\subsection{AMR-to-Logic Translation}
\label{sec:translator_specification}

This section specifies the deterministic AMR-to-logic translation shared by
every evaluated symbolic decoder. We use
AMR~\cite{banarescu-etal-2013-abstract}, PropBank for numbered semantic
roles~\cite{kingsbury-palmer-2002-treebank}, and the open-source Penman library
for PENMAN decoding~\cite{goodman-2020-penman}. Given one statement
\(x\), the fixed parser--translator pipeline either returns a Boolean formula,
its active atoms, and their base verbalizations, or fails without emitting a
translator frame.
A statement is one translator input unit and may contain multiple orthographic
sentences; parser \texttt{multi-sentence} branches are compiled conjunctively.

\begin{definition}[AMR Translator Output]
Fix an AMR parser configuration \(\vartheta\), and let
\(\mathcal T_{\mathrm{AMR}}^{\vartheta}\) be the resulting partial translator.
The parser and decoder first convert \(x\) into a normalized AMR graph, with
supported inverse roles mapped to their base-role directions. For every
successfully translated statement, let
\((a_{x,1},\ldots,a_{x,n_x})\) be the deterministic active-atom order. The
translator returns
\[
\mathfrak F_x=
\left\langle
\Phi_x,
\left(
\left\langle
\operatorname{id}(a_{x,i}),
\theta_x(a_{x,i}),
v_x(a_{x,i})
\right\rangle
\right)_{i=1}^{n_x}
\right\rangle.
\]
Let
\[
\mathcal A_x=\{a_{x,1},\ldots,a_{x,n_x}\}
\]
denote the corresponding active atom set. Here \(\Phi_x\) is the emitted
formula abstract syntax tree (AST), \(\theta_x(a)\) is the structured atom
expression, and \(v_x(a)\)
is its nonempty unsigned base verbalization. The translator does not add a
terminal sentence delimiter, and negation is represented in \(\Phi_x\), not
inserted into \(v_x(a)\). The downstream surface adapter strips surrounding
whitespace from \(v_x(a)\) and appends the string \texttt{.}, yielding the
main-paper surface \(\mathcal V_x(a)\). This formatting step is outside
\(\mathcal T_{\mathrm{AMR}}^{\vartheta}\). The solver-facing representation
used in the main paper is therefore
\[
\mathsf{Rep}(x)=\langle\mathcal A_x,\Phi_x,\mathcal V_x\rangle.
\]
If parsing, AMR decoding, compilation, or output validation fails,
\(\mathcal T_{\mathrm{AMR}}^{\vartheta}(x)\) is undefined; no fallback output
is substituted.
\end{definition}

\begin{example}[Running translator input]
\label{ex:translator_running_input}
For \(x=\textit{A careful student reads a book}\), suppose the fixed parser
returns
\begin{verbatim}
(r / read-01
   :ARG0 (s / student
      :mod (c / careful))
   :ARG1 (b / book))
\end{verbatim}
The graph determines the record partition, atom inventory, and Boolean
formula. Numbered-role verbalizations additionally use the pinned PropBank
role index. The symbols \(a_i\) used below are expository labels and need not
match emitted \texttt{x}\(i\) identifiers or their order.
\end{example}

\subsection{Records, Atoms, and Verbalizations}

\begin{definition}[Normalized AMR Records and Endpoint Descriptors]
Write the normalized graph for statement \(x\) as
\[
G_x=(\mathcal N_x,\mathcal E_x^{\mathrm N},
     \mathcal E_x^{\mathrm L},c_x,t_x),
\]
where \(\mathcal N_x\) is the node set,
\(\mathcal E_x^{\mathrm N}\) contains node-valued role records,
\(\mathcal E_x^{\mathrm L}\) contains literal-valued attribute records,
\(c_x\) maps nodes to concepts, and \(t_x\) is the parser top. Every
role or attribute occurrence receives a deterministic within-graph record
identifier; the compiler does not reconstruct occurrences from surface text.
Let
\(\mathcal E_x=\mathcal E_x^{\mathrm N}\biguplus
\mathcal E_x^{\mathrm L}\), where \(\biguplus\) denotes disjoint union.
Normalization deterministically partitions the records as
\[
\mathcal E_x=
\mathcal E_x^{\mathrm{sem}}\biguplus
\mathcal E_x^{\mathrm{str}}\biguplus
\mathcal E_x^{\mathrm{meta}}.
\]
Semantic records can produce dyadic records or contribute to triples.
Structural records determine Boolean structure. Metadata records may refine
endpoint identity or lexical realization; metadata used by neither remains
construction-only. Table~\ref{tab:record_partition} gives the principal cases.
List indices and descriptor subtrees are metadata under the same partition.
The only metadata role that may emit an active atom is the special active
\texttt{:mode} case; a mode attached to a structural connective remains
construction metadata. Metadata may affect a descriptor, but it does not
itself create a separate proposition.

Endpoint identity includes its concept, normalized name, and descriptor
metadata. Define
\[
\delta_x(u)=
\begin{cases}
\delta_x^{\mathrm N}(u),&u\in\mathcal N_x,\\
\delta_x^{\mathrm L}(u),&u\text{ is a literal},
\end{cases}
\]
where
\[
\begin{aligned}
\delta_x^{\mathrm N}(u)
  &=\operatorname{NodeDesc}\!\left(
    c_x(u),\operatorname{Name}_x(u),m_x(u)\right),\\
\delta_x^{\mathrm L}(u)
  &=\operatorname{LitDesc}\!\left(
    \operatorname{raw}(u),\operatorname{val}(u),
    \operatorname{type}(u)\right).
\end{aligned}
\]
Here \(\operatorname{Name}_x(u)\) is the normalized name, if present. In the
descriptor for
\(u\), the direct metadata list \(m_x(u)\) excludes \texttt{:name}, \texttt{name-part},
and \texttt{:wiki} records; node-valued metadata targets are serialized
recursively. The normalized name is already represented by
\(\operatorname{Name}_x(u)\), while
direct \texttt{:wiki} metadata is retained only for construction. For
literals, the raw PENMAN token, its evaluated value, and its recovered type
are retained. Parser node and record identifiers are source-record
identifiers and are not inserted into \(\delta_x\).
\end{definition}

\begin{table*}[t]
\centering
\begin{tabular}{@{}p{0.28\textwidth}p{0.38\textwidth}p{0.22\textwidth}@{}}
\toprule
\textbf{AMR record} & \textbf{Compiler action} & \textbf{Formula effect} \\
\midrule
\texttt{and}/\texttt{or} with \texttt{:op}\(i\) &
ordered branch traversal & \(\land\) / \(\lor\) \\
\texttt{multi-sentence} with \texttt{:snt}\(i\) or \texttt{:rel} &
ordered sentence traversal & \(\land\) \\
\texttt{:polarity -} & local scope marker & \(\lnot\) \\
\texttt{:condition} & antecedent edge & \(\to\) \\
\texttt{:name}, \texttt{:quant}, \texttt{:unit}, and date/value fields &
endpoint identity or lexical realization & no separate atom \\
\texttt{:wiki} & construction-only metadata & no separate atom \\
\bottomrule
\end{tabular}
\caption{Principal structural and metadata cases. Records not assigned to a
structural or metadata category are semantic records.}
\label{tab:record_partition}
\end{table*}

\begin{example}[Record partition and endpoint identity]
In Example~\ref{ex:translator_running_input}, the role occurrences
\texttt{:ARG0(r,s)}, \texttt{:ARG1(r,b)}, and \texttt{:mod(s,c)} belong to
\(\mathcal E_x^{\mathrm{sem}}\). There is no connective or scope record, so
\(\mathcal E_x^{\mathrm{str}}=\varnothing\). If the book node additionally
contained \texttt{:quant 1}, that occurrence would belong to
\(\mathcal E_x^{\mathrm{meta}}\): it would refine \(\delta_x(b)\) without
becoming an independent proposition.
\end{example}

\subsubsection{Active Atoms}

The concrete AMR role labels \(r,r_a,r_b\) below instantiate the main-paper
role parameters \(\kappa,\kappa_a,\kappa_b\), respectively; the predicate
occurrence \(p\) corresponds to \(e\) in the main paper's compact atom
notation.

\begin{definition}[Role-Aware Dyads and Same-Event Triples]
For a semantic record \((p,\texttt{:}r,u)\), where \(u\) may be a node or
literal, the compiler first creates the role-aware dyadic record
\[
\operatorname{Dya}_{r}\bigl(\delta_x(p),\delta_x(u)\bigr).
\]
Its payload records the role, both typed descriptors, and whether the
endpoints are the same node, distinct nodes, or a node and a literal. The
originating record identifier is stored separately as provenance.

Two node-valued role records can be compressed into one triple only when they
share the same predicate occurrence \(p\). Let
\[
\begin{aligned}
\mathcal C_{\mathrm{comp}}=\{&
\texttt{ARG0},\ldots,\texttt{ARG4},\\
&\texttt{accompanier},\texttt{beneficiary},\texttt{cause},\\
&\texttt{destination},\texttt{direction},\texttt{duration},\\
&\texttt{extent},\texttt{instrument},\texttt{location},\\
&\texttt{manner},\texttt{medium},\texttt{path},\\
&\texttt{purpose},\texttt{source},\texttt{time},\texttt{topic}\}.
\end{aligned}
\]
Among the core roles present on \(p\), the compiler chooses the first available
anchor in the fixed order \(\texttt{ARG0},\ldots,\texttt{ARG4}\). If
\((p,\texttt{:}r_a,u)\) is the anchor edge and
\((p,\texttt{:}r_b,v)\) is another edge with
\(r_b\in\mathcal C_{\mathrm{comp}}\), they produce
\[
\operatorname{Tri}_{r_a,r_b}
  \bigl(\delta_x(u),\delta_x(p),\delta_x(v)\bigr).
\]
This triple records that, in one occurrence of predicate \(p\), endpoints
\(u\) and \(v\) fill roles \(r_a\) and \(r_b\). When \texttt{ARG0} is absent,
a later core role can
anchor the event; in particular, \texttt{ARG1+ARG2} is legal. Cross-event
paths are never merged.

For each non-anchor role \(r_b\in\mathcal C_{\mathrm{comp}}\), the compiler
forms a separate triple with the selected anchor role. An event with
\texttt{ARG0}, \texttt{ARG1}, and \texttt{ARG2} yields
\(\operatorname{Tri}_{\mathrm{ARG0},\mathrm{ARG1}}\) and
\(\operatorname{Tri}_{\mathrm{ARG0},\mathrm{ARG2}}\); when \texttt{ARG0} is
available, the compiler does not additionally form an
\texttt{ARG1+ARG2} triple. Component dyads remain in the internal record
inventory, but once absorbed they are not active formula atoms. If exactly one
endpoint of a candidate triple is a structural connective, the compiler
recursively projects that endpoint to its leaf branches, creates one triple per
leaf, and preserves the nested connective topology in the formula. If the two
endpoints are independently coordinated, the compiler forms no Cartesian
product; the corresponding relations remain projected dyads.

Each distinct unconsumed semantic record whose source is not a structural connective and
whose target is not a structural connective produces one active dyad. If its
target is a structural connective, the relation is projected to one dyad per
leaf branch and the formula preserves the connective topology. Duplicate
semantic occurrences are represented once, with their record identifiers
retained together as internal provenance. A relation whose source is a
connective is represented by one opaque atom rather than being distributed
over its branches. A unary carrier is retained only when a concept would
otherwise have no active proposition or a negative node requires a nonempty
local scope.

Let \(\mathcal D_x^{\mathrm{rec}}\) be the dyadic-record inventory,
let \(\mathcal C_x\subseteq\mathcal D_x^{\mathrm{rec}}\) contain the component
dyads consumed by successful same-event composition, and let
\(\operatorname{atom}_{\mathrm D}(d)\) denote the dyadic proposition represented
by record \(d\). The provisional active inventory is
\[
\widetilde{\mathcal A}_x=
\widetilde{\mathcal A}_x^{\mathrm U}\biguplus
\widetilde{\mathcal A}_x^{\mathrm D}\biguplus
\widetilde{\mathcal A}_x^{\mathrm T}\biguplus
\widetilde{\mathcal A}_x^{\mathrm O},
\]
where the four components contain unary carriers, active unconsumed dyads,
same-event triples, and opaque atoms. The active dyads may also include the
special \texttt{:mode} case described above. Consumed dyads
satisfy
\[
\{\operatorname{atom}_{\mathrm D}(d):d\in\mathcal C_x\}
\cap\widetilde{\mathcal A}_x=\varnothing.
\]
They remain internal provenance records but cannot also be scored as active
atoms.
\end{definition}

\begin{example}[Triple composition and a residual dyad]
\label{ex:amr_atom_construction}
For the graph in Example~\ref{ex:translator_running_input}, the same-event
records \texttt{:ARG0(r,s)} and \texttt{:ARG1(r,b)} compose into
\[
a_1=\operatorname{Tri}_{\mathrm{ARG0},\mathrm{ARG1}}
 (\textit{student},\textit{read-01},\textit{book}).
\]
For readability, examples abbreviate typed endpoint descriptors by their
lexical heads; the compiler retains the full \(\delta_x\) values. The modifier
is not part of the event-role pair and remains
\[
a_2=\operatorname{Dya}_{\mathrm{mod}}
 (\textit{student},\textit{careful}).
\]
Thus \(\widetilde{\mathcal A}_x=\{a_1,a_2\}\). The \texttt{ARG0} and \texttt{ARG1}
dyads remain provenance records but are not additional active atoms.
\end{example}

\subsubsection{Atom Identity, Provenance, and Verbalization}

\begin{definition}[Atom Identity, Provenance, and Verbalization]
For every provisional atom \(a\), \(\theta_x(a)\) is its structured
representation: atom kind, role or role pair, ordered endpoint descriptors,
and, where applicable, a coreference flag. Its canonical key is
\[
\operatorname{key}_x(a)=\operatorname{Canon}\!\left(\theta_x(a)\right),
\]
where \(\operatorname{Canon}\) is a deterministic serialization independent
of parser variable names. In contrast,
\(\operatorname{prov}_x(a)\subseteq\mathcal E_x\) records the graph records
used to construct \(a\), and
\(\operatorname{owner}_x(a)\in\mathcal N_x\) is the graph node whose local
formula body contains the leaf for \(a\). Metadata may affect
\(\theta_x(a)\) or \(v_x(a)\) without appearing in
\(\operatorname{prov}_x(a)\); unary carriers may have empty provenance. These
fields are compiler-internal and are distinct from the emitted verbalization
\(v_x(a)\).

For a dyad \(\operatorname{Dya}_{r}(c,d)\) and a triple
\(\operatorname{Tri}_{r_a,r_b}(c,p,d)\), let \(C=\ell(c)\),
\(P=\ell(p)\), and \(D=\ell(d)\). The deterministic unsigned lexical form
\(\ell\) removes a predicate-sense suffix and realizes names, quantities, and
descriptor metadata from the AMR payload; it never consults the original
sentence. For an endpoint descriptor \(z\) and numbered role \(r\), the
resolver \(\chi(z,r)\) consults the predicate-sense entry stored in \(z\) and
returns a semantic relation class from the pinned PropBank roleset, or
\(\bot\) when the roleset or role is unavailable or unresolved. Non-numbered
roles are realized directly from their AMR role labels. Write
\(v_{\mathrm D}^{r}\) and
\(v_{\mathrm T}^{r_a,r_b}\) for the base verbalizations of the
corresponding dyad and triple. Subscripts \(\mathrm D\) and \(\mathrm T\)
denote dyadic and triple forms, respectively. Let \(\mathsf{Lex}^{\mathrm D}\) and
\(\mathsf{Lex}^{\mathrm T}\) denote the fixed dyadic and triple realization
maps specified below. Then
\[
\begin{aligned}
\operatorname{rel}_{\mathrm D}&=\chi(c,r),\\
\operatorname{rel}_{\mathrm T}&=\chi(p,r_b),\\
\operatorname{sig}_{\mathrm D}&=(r,\operatorname{rel}_{\mathrm D}),\\
\operatorname{sig}_{\mathrm T}&=(r_a,r_b,\operatorname{rel}_{\mathrm T}),\\
f_{\mathrm D}^{r}(c,d)&=
\begin{cases}
\operatorname{norm}(D\ C),&r=\texttt{ARG0},\\
\operatorname{norm}(C\ D),&\text{otherwise},
\end{cases}\\
f_{\mathrm T}(c,p,d)&=\operatorname{norm}(C\ P\ D),\\
v_{\mathrm D}^{r}(c,d)
  &=\mathsf{Lex}^{\mathrm D}_{\operatorname{sig}_{\mathrm D}}(C,D),\\
v_{\mathrm T}^{r_a,r_b}(c,p,d)
  &=\mathsf{Lex}^{\mathrm T}_{\operatorname{sig}_{\mathrm T}}(C,P,D).
\end{aligned}
\]
Here \(\operatorname{norm}\) joins nonempty lexical fields with single spaces.
If the relevant resolver output is \(\bot\), or a supported join has no
resolved relation template, \(\mathsf{Lex}^{\mathrm D}\) returns
the ordered fallback \(f_{\mathrm D}^{r}\), and
\(\mathsf{Lex}^{\mathrm T}\) returns \(f_{\mathrm T}\). An opaque
atom uses the dyadic template for its retained role. For a unary carrier with
concept \(c\), lexical form \(C=\ell(c)\), and optional quantity \(\nu\), define
\[
\mathsf{Lex}^{\mathrm U}(C;c,\nu)=
\begin{cases}
C\ \textit{occurs},&c\text{ is sense-tagged},\\
C\ \textit{exist},&\nu\text{ is present and }\nu\neq 1,\\
C\ \textit{exists},&\text{otherwise}.
\end{cases}
\]

The selected template depends only on the active atom kind, ordered terms,
role or role pair, and, for numbered roles, the specific predicate-sense entry in
PropBank. Tables~\ref{tab:core_dyad_templates},
\ref{tab:nonnumbered_dyad_templates}, and \ref{tab:triple_templates}
summarize the surface templates applied after role resolution. The fixed
renderer additionally applies the specified PropBank-description refinements,
property-predicate cases, and rule-based inflections. These affect only
\(v_x\), not structured atom identity or the Boolean formula. A
listed metadata or connective role does not itself activate a proposition;
activation is fixed by the record partition and atom construction above. In
the tables,
\(\operatorname{pp}(\cdot)\) is the fixed rule-based
past-participle inflector.

Surface realization leaves \(\theta_x\), \(\operatorname{key}_x\),
\(\operatorname{owner}_x\), \(\operatorname{prov}_x\), and the formula
abstract syntax tree unchanged.
Negation is carried by \(\Phi_x\) and is never inserted into the
verbalization.
\end{definition}

\begin{example}[Atom identity and verbalization]
For the atoms in Example~\ref{ex:amr_atom_construction}, the structured
representation of
\(a_1\) retains the predicate sense, role pair, and endpoint descriptors,
while its verbalization omits the sense suffix:
\[
\begin{aligned}
\theta_x(a_1)
  &=\operatorname{Tri}_{\mathrm{ARG0},\mathrm{ARG1}}
    (\textit{student},\textit{read-01},\textit{book}),\\
\operatorname{key}_x(a_1)
  &=\operatorname{Canon}\!\left(\theta_x(a_1)\right),\\
v_x(a_1)&=\textit{``student read book''},\\
v_x(a_2)&=\textit{``careful student''}.
\end{aligned}
\]
Thus the structured expression preserves \texttt{read-01}, whereas the
verbalization uses \textit{read}. Formula construction yields
\(\Phi_x=a_1\land a_2\).
\end{example}

\begin{table*}[p]
\centering
\begin{tabular}{@{}p{0.28\textwidth}p{0.64\textwidth}@{}}
\toprule
\multicolumn{2}{c}{\textbf{(a) PropBank function-tag resolution}} \\
\midrule
\textbf{PropBank tag} & \textbf{Resolved relation class} \\
\midrule
\texttt{PAG}, \texttt{CAU}, \texttt{SRC}, \texttt{EXT},
\texttt{PRP}, \texttt{TMP}
& agent, cause, source, extent, purpose, and time, respectively \\
\texttt{PPT}
& refine to agent, property, instrument, accompaniment, path, source,
location, topic, extent, attribute, active theme, or state; otherwise patient \\
\texttt{GOL}
& refine to beneficiary, result, location, or instrument; otherwise goal \\
\texttt{LOC}
& distinguish path, source, and goal; otherwise location \\
\texttt{DIR}
& distinguish patient, source, goal, path, location, and beneficiary;
otherwise direction \\
\texttt{MNR} & refine to instrument or accompaniment; otherwise manner \\
\texttt{COM} & distinguish opponent and companion \\
\texttt{PRD} & patient for \texttt{ARG1}; result otherwise \\
unavailable, missing, \texttt{VSP}, or unrecognized
& return \(\bot\) and use the ordered fallback \\
\midrule
\multicolumn{2}{c}{\textbf{(b) Numbered-role surface templates}} \\
\midrule
\textbf{Resolved relation class} &
\textbf{\(v(\operatorname{Dya}_{r}(c,d))\)} \\
\midrule
agent; active theme & \(D\ C\) \\
patient & \(D\ \textit{is}\ \operatorname{pp}(C)\) \\
goal & \(C\ \textit{is directed to}\ D\) \\
location & \(C\ \textit{occurs at}\ D\) \\
path & \(C\ \textit{occurs along}\ D\) \\
direction & \(C\ \textit{proceeds toward}\ D\) \\
source & \(C\ \textit{originates from}\ D\) \\
instrument & \(C\ \textit{uses}\ D\) \\
manner & \(C\ \textit{occurs by}\ D\) \\
companion; accompaniment & \(C\ \textit{occurs with}\ D\) \\
opponent & \(C\ \textit{occurs against}\ D\) \\
beneficiary; purpose & \(C\ \textit{is for}\ D\) \\
result & \(C\ \textit{results in}\ D\) \\
extent & \(C\ \textit{has extent}\ D\) \\
cause & \(C\ \textit{is caused by}\ D\) \\
topic & \(C\ \textit{is about}\ D\) \\
time & \(C\ \textit{occurs during}\ D\) \\
property & \(D\ \textit{is}\ C\) \\
state & \(D\ \textit{is in state}\ C\) \\
attribute & \(C\ \textit{has attribute}\ D\) \\
unresolved & \(f_{\mathrm D}^{r}\) \\
\bottomrule
\end{tabular}
\caption{PropBank-numbered dyadic verbalization. Panel (a) resolves
PropBank function tags to relation classes; panel (b) maps those classes to
surface templates. Here \(C\) is the predicate surface and \(D\) is the
argument surface. Relation classes that share a row use the same template.}
\label{tab:core_dyad_templates}
\end{table*}

\begin{table*}[p]
\centering
\begin{tabular}{@{}p{0.32\textwidth}p{0.60\textwidth}@{}}
\toprule
\textbf{AMR role or atom kind} & \textbf{Verbalization template} \\
\midrule
\texttt{purpose}, \texttt{beneficiary} & \(C\ \textit{is for}\ D\) \\
\texttt{time}, \texttt{location} & \(C\ \textit{occurs at}\ D\) \\
\texttt{direction} & \(C\ \textit{proceeds toward}\ D\) \\
\texttt{domain} & \(D\ \textit{is}\ C\) \\
\texttt{mod} & \(D\ C\) \\
\texttt{manner} & \(C\ \textit{occurs in manner}\ D\) \\
\texttt{poss} & \(D\textit{'s}\ C\) \\
\texttt{poss-of} & \(C\textit{'s}\ D\) \\
\texttt{topic} & \(C\ \textit{is about}\ D\) \\
\texttt{part}, \texttt{subevent-of} & \(D\ \textit{is part of}\ C\) \\
\texttt{part-of}, \texttt{subevent} & \(C\ \textit{is part of}\ D\) \\
\texttt{consist} & \(D\ \textit{consists of}\ C\) \\
\texttt{consist-of} & \(C\ \textit{consists of}\ D\) \\
\texttt{location-of} & \(D\ \textit{occurs at}\ C\) \\
\texttt{dayperiod} & \(C\ \textit{occurs during}\ D\) \\
\texttt{destination} & \(C\ \textit{proceeds to}\ D\) \\
\texttt{source} & \(C\ \textit{originates from}\ D\) \\
\texttt{instrument} & \(C\ \textit{uses}\ D\) \\
\texttt{accompanier} & \(C\ \textit{occurs with}\ D\) \\
\texttt{path} & \(C\ \textit{proceeds along}\ D\) \\
\texttt{medium} & \(C\ \textit{occurs via}\ D\) \\
\texttt{cause} & \(C\ \textit{is caused by}\ D\) \\
\texttt{concession} & \(C\ \textit{occurs despite}\ D\) \\
\texttt{duration} & \(C\ \textit{lasts for}\ D\) \\
\texttt{degree}, \texttt{age} &
\(C\ \textit{has degree}\ D\); \(C\ \textit{has age}\ D\) \\
\texttt{frequency} & \(C\ \textit{has frequency}\ D\) \\
\texttt{extent}, \texttt{example} &
\(C\ \textit{has extent}\ D\); \(C\ \textit{has example}\ D\) \\
\texttt{:prep-}\(q\) & \(C\ q\ D\) \\
\texttt{:op}\(i\) & \(C\ D\) \\
special \texttt{mode} & \(C\ \textit{has}\ D\ \textit{mode}\) \\
other non-numbered role & \(f_{\mathrm D}^{r}\) \\
unary predicate / entity &
\(C\ \textit{occurs}\) / \(C\ \textit{exists}\)
(\textit{exist} when \(\nu\neq 1\)) \\
\bottomrule
\end{tabular}
\caption{Surface templates for non-numbered dyads and unary carriers. For
\texttt{:prep-}\(q\), \(q\) is the role suffix with hyphens replaced by
spaces.}
\label{tab:nonnumbered_dyad_templates}
\end{table*}

\begin{table*}[p]
\centering
\begin{tabular}{@{}p{0.39\textwidth}p{0.53\textwidth}@{}}
\toprule
\textbf{Join or resolved relation class} &
\textbf{\(v(\operatorname{Tri}_{r_a,r_b}(c,p,d))\)} \\
\midrule
\texttt{ARG0+ARG1} & \(C\ P\ D\) \\
\texttt{accompanier}; \texttt{beneficiary}/\texttt{purpose} &
\(C\ P\ \textit{with}\ D\); \(C\ P\ \textit{for}\ D\) \\
\texttt{cause}; \texttt{destination}; \texttt{direction} &
\(C\ P\ \textit{because of}\ D\); \(C\ P\ \textit{to}\ D\);
\(C\ P\ \textit{toward}\ D\) \\
\texttt{duration}; \texttt{extent} &
\(C\ P\ \textit{for}\ D\); \(C\ P\ \textit{by}\ D\) \\
\texttt{instrument}; \texttt{location}; \texttt{manner} &
\(C\ P\ \textit{using}\ D\); \(C\ P\ \textit{at}\ D\);
\(C\ P\ \textit{by}\ D\) \\
\texttt{medium}; \texttt{path}; \texttt{source} &
\(C\ P\ \textit{via}\ D\); \(C\ P\ \textit{along}\ D\);
\(C\ P\ \textit{from}\ D\) \\
\texttt{time}; \texttt{topic} &
\(C\ P\ \textit{during}\ D\); \(C\ P\ \textit{about}\ D\) \\
goal; location; path; direction; source &
\(C\ P\ \textit{to/at/along/toward/from}\ D\) \\
instrument; manner & \(C\ P\ \textit{using/by}\ D\) \\
companion or accompaniment; opponent &
\(C\ P\ \textit{with/against}\ D\) \\
beneficiary or purpose; result &
\(C\ P\ \textit{for}\ D\); \(C\ P\ D\) \\
extent; cause; topic; time &
\(C\ P\ \textit{by/because of/about/during}\ D\) \\
property, state, or attribute & \(C\ P\ \textit{as}\ D\) \\
agent, patient, or active theme as a secondary role &
\(f_{\mathrm T}\) \\
unresolved numbered role or unsupported join & \(f_{\mathrm T}\) \\
\bottomrule
\end{tabular}
\caption{Surface templates for same-event triple atoms. Monospaced entries are
AMR roles or joins; roman entries are PropBank-resolved classes or fallbacks.
Here \(C\), \(P\), and \(D\) are the anchor, predicate, and non-anchor surfaces.
Explicit adjunct roles precede PropBank classes. Slash-separated prepositions
map in order to the listed classes; no slash is emitted.}
\label{tab:triple_templates}
\end{table*}
\subsection{Formula Construction, Polarity, and Validation}

\begin{definition}[Recursive Formula Construction]
\label{def:recursive_formula_construction}
For specification, let
\[
\mathsf{Trav}_x=(\operatorname{owner}_x,\operatorname{Ch}_x,
       \operatorname{Op}_x,
       \operatorname{Cond}_x,\mathcal Q_x^{0})
\]
summarize the compiler's formula traversal. Here \(\operatorname{owner}_x\)
assigns every provisional atom to one formula owner;
\(\operatorname{Ch}_x(v)\) is the ordered list of semantic children
reached from \(v\) through normalized forward or inverse-tree records;
\(\operatorname{Op}_x(v)\) contains ordered connective branches; and
\(\operatorname{Cond}_x(v)\) contains condition targets. Condition and
connective records are handled separately and are excluded from ordinary child
traversal. Metadata and other structural records are excluded from ordinary
child traversal. A semantic back-edge to an already open node is also not
traversed, because the corresponding relation is already represented by a
dyad or triple.

The primary roots \(\mathcal Q_x^{0}\) contain the parser top and any
incoming-free roots of disconnected graph components. After compiling them,
if a provisional atom is absent from the formula, its owner is added as an
additional coverage root in \(\mathcal Q_x^{\mathrm{cov}}\). Coverage continues
until every provisional atom is represented; failure to add a new leaf is an
invalid translation.
Let
\[
\mathcal Q_x=\mathcal Q_x^{0}\biguplus
\mathcal Q_x^{\mathrm{cov}}.
\]
For a provisional atom \(a\), let \(\operatorname{id}(a)\) be its assigned
identifier. For a formula \(\varphi\), let
\(\operatorname{LeafIds}(\varphi)\) be the set of identifiers on its atom
leaves.

For node \(v\), let \(\operatorname{Local}_x(v)\) be the conjunction of
provisional atoms owned by \(v\), and define
\[
H(v)=\bigwedge_{u\in\operatorname{Ch}_x(v)}F(u),
\qquad
B_0(v)=\operatorname{Local}_x(v)\land H(v).
\]
We take \(\bigwedge\varnothing=\top\) and
\(\bigvee\varnothing=\bot\). Let
\(\mathcal C_{\land}=\{\texttt{and},\texttt{multi-sentence}\}\). For a
connective node with ordered branch roots \(u_i\), define
\[
\operatorname{Conn}_x(v)=
\begin{cases}
\bigwedge_i F(u_i),&c_x(v)\in\mathcal C_{\land},\\
\bigvee_i F(u_i),&c_x(v)=\texttt{or}.
\end{cases}
\]
The targets \(u_i\) arise from \texttt{:op}\(i\), \texttt{:snt}\(i\), or a
multi-sentence \texttt{:rel} edge. The local body is
\[
B(v)=
\begin{cases}
B_0(v)\land\operatorname{Conn}_x(v),
  &\operatorname{Conn}_x(v)\text{ is defined},\\
B_0(v),&\text{otherwise}.
\end{cases}
\]
Node polarity is applied first:
\[
N(v)=
\begin{cases}
\lnot B(v),&v\text{ has }\texttt{:polarity -},\\
B(v),&\text{otherwise},
\end{cases}
\]
Conditions are then applied to the resulting local body. If
\(\operatorname{Cond}_x(v)=\{q_1,\ldots,q_k\}\), define
\[
F(v)=
\begin{cases}
\left(\bigwedge_j F(q_j)\right)\to N(v),
  &\operatorname{Cond}_x(v)\neq\varnothing,\\
N(v),&\text{otherwise}.
\end{cases}
\]
Thus polarity on a conditioned node negates its consequent, not the whole
implication. AMR reentrancy is not expanded a second time when it
points to an already open node; the corresponding relation is already present
in a dyad or triple. If a condition target is a proper ancestor already open on
the current traversal path, the compiler uses the conjunction of that target
node's locally owned atoms as a finite antecedent. A self-condition or an empty
local antecedent is invalid. A cycle formed entirely by structural connective
or multi-sentence branch edges is also invalid rather than repaired; a branch
already open only through ordinary semantic reentrancy is not expanded again.

The formula before participant-local polarity projection is
\[
\Phi_x^{0}=\bigwedge_{q\in\mathcal Q_x}F(q).
\]
Coverage-root construction guarantees
\[
\operatorname{LeafIds}(\Phi_x^{0})
=\{\operatorname{id}(a):a\in\widetilde{\mathcal A}_x\}.
\]
\end{definition}

\begin{example}[Coordination and condition]
Suppose the parser represents
\textit{The committee chooses tea, or both coffee and cake} with an outer
\texttt{or} branch whose second child is an \texttt{and}. Let
\(a_t,a_c,a_k\) denote the three same-event choice atoms. Then
Definition~\ref{def:recursive_formula_construction} gives
\[
\Phi_x=a_t\lor(a_c\land a_k).
\]
For \texttt{go-01 :ARG0 child :condition rain-01}, let \(a_g\) be the
\texttt{ARG0} dyad for the going event and \(a_r\) the unary carrier for the
otherwise atomless condition node. Since the condition forms the antecedent
and the local body forms the consequent, the result is
\(\Phi_x=a_r\to a_g\).
\end{example}

\subsubsection{Polarity, Branch Isolation, and Validation}

\begin{definition}[Participant-Local Polarity]
Node polarity on an event or proposition is handled by \(N(v)\) above. An
explicit negative participant requires a narrower operation: only relation
atoms containing that participant are negated. Here a participant endpoint is
a triple subject or object, or the node-valued target of a dyad; unary and
opaque atoms are not projected by this rule. Let
\[
\mathcal P_x=
\left\{\operatorname{id}(a)\ \middle|\
\begin{array}{l}
a\in\widetilde{\mathcal A}_x^{\mathrm D}
   \cup\widetilde{\mathcal A}_x^{\mathrm T},\\
a\text{ contains an explicitly negative}\\
\text{participant endpoint}
\end{array}
\right\}.
\]
For parity bit \(b\in\{0,1\}\), define \(\Pi_{\mathcal P_x}\) recursively by
\[
\begin{aligned}
\Pi_{\mathcal P_x}(a;b)
  &=\begin{cases}
      \lnot a,&\operatorname{id}(a)\in\mathcal P_x\text{ and }b=0,\\
      a,&\text{otherwise},
    \end{cases}\\
\Pi_{\mathcal P_x}(\lnot\varphi;b)
  &=\lnot\Pi_{\mathcal P_x}(\varphi;1-b),\\
\Pi_{\mathcal P_x}
  \left(\bigcirc_{i=1}^{k}\varphi_i;b\right)
  &=\bigcirc_{i=1}^{k}\Pi_{\mathcal P_x}(\varphi_i;b),
    \qquad \bigcirc\in\{\land,\lor\},\\
\Pi_{\mathcal P_x}(\varphi\to\psi;b)
  &=\Pi_{\mathcal P_x}(\varphi;b)
    \to\Pi_{\mathcal P_x}(\psi;b).
\end{aligned}
\]
Constants are unchanged. Let
\[
\widetilde{\Phi}_x=\Pi_{\mathcal P_x}(\Phi_x^{0};0)
\]
be the formula before release finalization. The emitted formula and atom
inventory are
\[
\begin{aligned}
\Phi_x&=\operatorname{ConstNorm}(\widetilde{\Phi}_x),\\
\mathcal A_x&=
\left\{a\in\widetilde{\mathcal A}_x:
\operatorname{id}(a)\in\operatorname{LeafIds}(\Phi_x)\right\}.
\end{aligned}
\]
Here \(\operatorname{ConstNorm}\) applies the usual truth-preserving identities
for \(\top\), \(\bot\), negation, conjunction, disjunction, and implication.
Removed atoms are not emitted. The parity guard prevents a second negation
when the same leaf occurrence is
already under logical negation. Projection preserves every Boolean operator,
branch, atom owner, and provenance record; it neither applies De Morgan's law
nor moves participant polarity to the predicate.
\end{definition}

\begin{example}[Event and participant polarity]
For \textit{The teacher did not give the student a book}, the event has
\texttt{ARG0}, \texttt{ARG1}, \texttt{ARG2}, and \texttt{:polarity -}. Its
two active atoms are
\[
\begin{aligned}
g_1&=\operatorname{Tri}_{\mathrm{ARG0},\mathrm{ARG1}}
 (\textit{teacher},\textit{give-01},\textit{book}),\\
g_2&=\operatorname{Tri}_{\mathrm{ARG0},\mathrm{ARG2}}
 (\textit{teacher},\textit{give-01},\textit{student}).
\end{aligned}
\]
Event polarity preserves their shared parser scope:
\[
\Phi_x=\lnot(g_1\land g_2),
\]
not \((\lnot g_1)\land(\lnot g_2)\). By contrast, polarity on an explicit
participant is projected locally to each relation atom containing that
participant. For an AMR containing
\texttt{play-01 :ARG0 (child :polarity -) :location statue}, the output is the
single atom \(p_1\), with
\(v_x(p_1)=\textit{``child play at statue''}\), under formula
\(\lnot p_1\). This participant-only projection does not change the atom
owner, negate the predicate, apply De Morgan's law, or alter the surrounding
\texttt{and}/\texttt{or} branch.
\end{example}

\begin{definition}[Branch Isolation]
When compiling one explicit \texttt{and}/\texttt{or} branch, the compiler
excludes atoms anchored only to sibling events and blocks inverse traversal
into sibling roots. If a shared owner has several event anchors, the anchor
reachable only through the active branch is used in that branch's formula
view. A participant descriptor may be shared across branches, but each event
atom remains in the branch containing its event occurrence.
\end{definition}

\begin{definition}[Translator Validity]
The emitted frame contains only a formula AST and active atom records. Each
active atom has a unique nonempty identifier, a valid structured expression,
and a nonempty unsigned verbalization. The formula AST may use only
\texttt{atom}, \texttt{not}, \texttt{and}, \texttt{or}, and
\texttt{implies}. A \texttt{not} node has one argument, an \texttt{and} or
\texttt{or} node has at least two arguments, and an \texttt{implies} node has
an antecedent and a consequent. The formula must satisfy
\[
\operatorname{LeafIds}(\Phi_x)
=\{\operatorname{id}(a):a\in\mathcal A_x\}.
\]
No Boolean constant remains in the emitted formula.

Any parsing, AMR decoding, compilation, or output-validation failure yields an
invalid translation, including an empty active inventory or a formula--atom
closure mismatch. No fallback translator or semantic-repair rule is applied.
\end{definition}

\paragraph{Semantic boundary.}
The translator preserves explicit AMR coordination, conditions, and polarity,
but does not infer modal, factive, or discourse-level semantics.

\begin{example}[Isolating reentrant connective branches]
Suppose the parser represents
\textit{The student reads a book and writes an essay} with a shared
\texttt{student} node reentered by both event branches. Branch-local ownership
then yields
\[
\begin{aligned}
&\operatorname{Tri}_{\mathrm{ARG0},\mathrm{ARG1}}
(\textit{student},\textit{read-01},\textit{book})\\
&\qquad\land
\operatorname{Tri}_{\mathrm{ARG0},\mathrm{ARG1}}
(\textit{student},\textit{write-01},\textit{essay}).
\end{aligned}
\]
\end{example}
\section{Detailed Calculation for the Constructed Example}
\label{sec:supp_constructed_example}

Throughout Sections~\ref{sec:supp_constructed_example}--\ref{sec:audited_failure_trace},
candidate and active-link-set subscripts are suppressed when unambiguous. We
use the fixed main-paper settings \(W_{\max}=100\) and \(\epsilon=1\). Here
\(M^{\mathrm{base}}\) and \(M^{\mathrm{forced}}\) are the assignments returned
by solving the base and target-forced MaxSAT instances, and
\(\operatorname{SemCost}\) sums the weights of violated atom-link clauses in
\(\Omega^{\mathrm{sem}}\).

This section expands the constructed common-target example from the main
paper. It shows how each local link outcome induces a MaxSAT calculation and
how the resulting resistances are aggregated by \textsc{Top-Link} and PWAL.

\paragraph{(a) Fixed translations and atom meanings.}
\[
\begin{aligned}
\Phi_P&=g, & \Phi_A&=a_1\land a_2,\\
\Phi_{S_A}&=g\land a_1\land a_2,
& \Phi_{T_A}&=\Phi_{T_B}=b,\\
\Phi_B&=d_1\land d_2,
& \Phi_{S_B}&=g\land d_1\land d_2.
\end{aligned}
\]
\(g\): door is unlocked;\quad
\(a_1,d_1\): Maya enter office;\quad
\(a_2\): Maya remain at office;\quad
\(d_2\): Maya leave office;\quad
\(b\): Maya present at office.

\paragraph{(b) NLI outputs and induced semantic clauses.}
The retained alternatives are
\[
\begin{aligned}
A:\quad& a_1\!\to b\;(\texttt{Ent},p=0.989,w=98),\\
       & a_2\!\to b\;(\texttt{Ent},p=0.995,w=99);\\
B:\quad& d_1\!\to b\;(\texttt{Ent},p=0.989,w=98),\\
       & d_2\!\to b\;(\texttt{Con},p=0.668,w=66).
\end{aligned}
\]
They induce the weighted clauses
\[
\begin{gathered}
\bigl[\neg a_1\lor b\bigr]_{98},\quad
\bigl[\neg a_2\lor b\bigr]_{99},\\
\bigl[\neg d_1\lor b\bigr]_{98},\quad
\bigl[\neg d_2\lor\neg b\bigr]_{66}.
\end{gathered}
\]
The shared comparison \(g\!\to b\) is neutral with confidence \(0.982\) and
is discarded.
Each candidate has one unmatched target atom and two non-neutral alternatives.
PWAL therefore assigns probability \(1/3\) to no-link and to each listed
candidate-specific link. Confidence determines \(w(p)\), not the
local-outcome probability.

\paragraph{(c) MaxSAT calculation by local outcome.}
In this constructed example, each active-link source is forced true by its
candidate's hard source formula. With target-inertia weight \(\epsilon=1\),
all displayed semantic-link weights exceed the inertia weight.
The resistance score is
\[
\begin{aligned}
\Delta T&=\operatorname{SemCost}(M^{\mathrm{forced}})\\
&\quad-\operatorname{SemCost}(M^{\mathrm{base}}),\\
\rho&=\frac{\Delta T}{C_{\max}}-R_{\mathrm{sat}}.
\end{aligned}
\]
In the no-link case, the base and forced values of \(b\) are false and true,
and both semantic costs are zero. Under entailment they are both true and both
costs are zero. Under a contradiction of weight \(w\), they are false and true,
so the costs are \(0\) and \(w\). Consequently,
\[
\begin{array}{lccc}
\toprule
\text{Outcome} & \Delta T/C_{\max} & R_{\mathrm{sat}} & \rho\\
\midrule
\text{No-link} & 0 & 0 & 0\\
\text{Entailment} & 0 & 1 & -1\\
\text{Contradiction} & 1 & 0 & 1\\
\bottomrule
\end{array}
\]
In the no-link row, \(C_{\max}=|\mathcal A_T|W_{\max}=100\) is the
zero-link fallback. Inertia affects the two optima but is excluded from
\(\operatorname{SemCost}\).

\paragraph{(d) Instantiated worlds and pairwise decision.}
For each world \(\omega\), report the score tuple
\(\mathbf s(\omega)=(\Delta T,C_{\max},R_{\mathrm{sat}},\rho)\). The complete world spaces,
with probability \(1/3\) per world, are
\[
\begin{aligned}
A:\quad&\mathbf s(\omega_{A0})=(0,100,0,0),\\
       &\mathbf s(\omega_{A1})=(0,98,1,-1),\quad
        \mathbf s(\omega_{A2})=(0,99,1,-1);\\
B:\quad&\mathbf s(\omega_{B0})=(0,100,0,0),\\
       &\mathbf s(\omega_{B1})=(0,98,1,-1),\quad
        \mathbf s(\omega_{B2})=(66,66,0,1).
\end{aligned}
\]
The \(\omega_{A1},\omega_{A2},\omega_{B1},\omega_{B2}\) worlds activate
\(a_1\!\to b\), \(a_2\!\to b\),
\(d_1\!\to b\), and \(d_2\!\to b\), respectively. Aggregation gives
\[
\begin{aligned}
\textsc{Top-Link}:&\quad
L_A^{\mathrm{top}}=\{a_2\!\to b\},\\
&\quad L_B^{\mathrm{top}}=\{d_1\!\to b\},\\
&\quad \sigma_A^{\mathrm{top}}=\sigma_B^{\mathrm{top}}=-1
\Longrightarrow \textbf{Tie};\\
\text{PWAL}:&\quad
\mu_A=\tfrac13(0-1-1)=-\tfrac23,\\
&\quad\mu_B=\tfrac13(0-1+1)=0,\\
&\quad \mu_A<\mu_B
\Longrightarrow \textbf{Candidate A}.
\end{aligned}
\]
\section{Detailed ARCT Success Trace}
\label{sec:audited_trace}

In this fully enumerable ARCT instance, \textsc{Top-Link} ties, whereas PWAL
selects the correct candidate for every seed 2026--2035 and under exact
marginalization. It is illustrative only and was not used for method or
setting selection.
Tables~\ref{tab:short_arct_input}--\ref{tab:short_arct_worlds_b} report the
input, retained links, and complete world aggregation.

\paragraph{(a) Input, fixed translations, and atom meanings.}

\begin{table*}[!t]
\centering
\begin{tabular}{@{}p{0.19\textwidth}p{0.75\textwidth}@{}}
\toprule
Premise & People learn a lot from comment sections. \\
Candidate A & too much learning hurts \\
Candidate B & learning never hurts \\
Claim & Comment sections have not failed \\
\bottomrule
\end{tabular}
\caption{ARCT success trace. Candidate B is gold.}
\label{tab:short_arct_input}
\end{table*}

The compiler produces
\[
\begin{aligned}
\Phi_{S_A}&=a_1\land a_2\land a_3\land a_4\land a_5\land a_6,\\
\Phi_{S_B}&=b_1\land b_2\land b_3\land\neg b_4,\\
\Phi_T&=\neg f\lor\neg c.
\end{aligned}
\]
The atom surfaces below are reproduced verbatim from the translator
output and were not manually edited. The atom inventory is
\[
\begin{aligned}
a_1,b_1&:\ \text{comment section},\\
a_2,b_2&:\ \text{person learn lot},\\
a_3,b_3&:\ \text{person learn toward section},\\
a_4&:\ \text{learn hurt},\quad
a_5:\ \text{hurt much},\quad a_6:\ \text{hurt too},\\
b_4&:\ \text{learn hurt during ever},\\
f&:\ \text{section fail},\quad c:\ \text{comment section}.
\end{aligned}
\]
The negative literal \(\neg b_4\) is the compiled meaning of ``learning never
hurts.'' Exact surface matching supplies the deterministic semantic links
\(a_1\leftrightarrow c\) and \(b_1\leftrightarrow c\), each with fixed
exact-link weight 100. Consequently only the unmatched target atom \(f\) is
sampled.

\paragraph{(b) NLI outputs, clause weights, and local outcomes.}

For an NLI confidence \(p\), the solver uses
\(w(p)=\max\{1,\lfloor100p\rfloor\}\). Entailment induces
\([\neg s\lor f]_{w(p)}\), whereas contradiction induces
\([\neg s\lor\neg f]_{w(p)}\).

\begin{table*}[!t]
\centering
\begin{tabular}{@{}cclcccl@{}}
\toprule
Candidate & Link & Source surface & Label & \(p\) & \(w(p)\) & Soft clause \\
\midrule
A & \(a_1\!\to f\) & comment section & \texttt{Con} & 0.848 & 84 & \([\neg a_1\lor\neg f]_{84}\) \\
A & \(a_2\!\to f\) & person learn lot & \texttt{Con} & 0.982 & 98 & \([\neg a_2\lor\neg f]_{98}\) \\
A & \(a_3\!\to f\) & person learn toward section & \texttt{Con} & 0.971 & 97 & \([\neg a_3\lor\neg f]_{97}\) \\
A & \(a_5\!\to f\) & hurt much & \texttt{Ent} & 0.421 & 42 & \([\neg a_5\lor f]_{42}\) \\
\midrule
B & \(b_1\!\to f\) & comment section & \texttt{Con} & 0.848 & 84 & \([\neg b_1\lor\neg f]_{84}\) \\
B & \(b_2\!\to f\) & person learn lot & \texttt{Con} & 0.982 & 98 & \([\neg b_2\lor\neg f]_{98}\) \\
B & \(b_3\!\to f\) & person learn toward section & \texttt{Con} & 0.971 & 97 & \([\neg b_3\lor\neg f]_{97}\) \\
B & \(b_4\!\to f\) & learn hurt during ever & \texttt{Ent} & 0.587 & 58 & \([\neg b_4\lor f]_{58}\) \\
\bottomrule
\end{tabular}
\caption{All retained candidate-specific NLI links. Each candidate's five
local outcomes---the four displayed links and no-link---have probability
\(1/5\).}
\label{tab:short_arct_links}
\end{table*}

Notice that the final B-link is built on the unsigned base atom \(b_4\), while
the source formula hard-enforces \(\neg b_4\). Its implication is therefore
satisfied through \(\neg b_4\) and changes neither the target-clause witness
ratio nor semantic tension in the corresponding world. This is why it is not
equivalent to A's positive ``hurt much'' entailment link.

\paragraph{(c) MaxSAT calculation and all enumerated worlds.}

The source formula is hard in both MaxSAT instances; \(\Phi_T\) is hard only
in the forced instance.  Unit target inertia participates in optimization but
is excluded from semantic cost.  For each world,
\[
\begin{aligned}
\Delta T&=\operatorname{SemCost}(M^{\mathrm{forced}})\\
&\quad-\operatorname{SemCost}(M^{\mathrm{base}}),\\
\rho&=\frac{\Delta T}{C_{\max}}-R_{\mathrm{sat}}.
\end{aligned}
\]
The exact match contributes 100 to \(C_{\max}\); an active NLI link contributes
its weight.  With no active NLI link, \(C_{\max}=100\).  The exact link makes
\(c\) true in the base optimum.  An active contradiction link also supports
\(\neg f\), thereby witnessing the sole target clause \(\neg f\lor\neg c\)
and giving \(R_{\mathrm{sat}}=1\). Tables~\ref{tab:short_arct_worlds_a} and
\ref{tab:short_arct_worlds_b} list the two complete five-world spaces. Every
world has probability \(1/5\); \(n_{2026}\) is its number of occurrences in
the fixed seed-2026 sample of 100 worlds for that candidate.

\begin{table}[t]
\centering
\begin{tabular}{@{}clrrc@{}}
\toprule
World & Link & \(n_{2026}\) & \((\Delta T,C_{\max})\) &
\((R_{\mathrm{sat}},\rho)\) \\
\midrule
A0 & no-link & 22 & \((0,100)\) & \((0,0)\) \\
A1 & \(a_1\!\to f\) & 17 & \((0,184)\) & \((1,-1)\) \\
A2 & \(a_2\!\to f\) & 17 & \((0,198)\) & \((1,-1)\) \\
A3 & \(a_3\!\to f\) & 17 & \((0,197)\) & \((1,-1)\) \\
A4 & \(a_5\!\to f\) & 27 & \((42,142)\) & \((0,0.295775)\) \\
\bottomrule
\end{tabular}
\caption{Candidate-A worlds and seed-2026 counts. Link labels and weights are
given in Table~\ref{tab:short_arct_links}.}
\label{tab:short_arct_worlds_a}
\end{table}

\begin{table}[t]
\centering
\begin{tabular}{@{}clrrc@{}}
\toprule
World & Link & \(n_{2026}\) & \((\Delta T,C_{\max})\) &
\((R_{\mathrm{sat}},\rho)\) \\
\midrule
B0 & no-link & 24 & \((0,100)\) & \((0,0)\) \\
B1 & \(b_1\!\to f\) & 29 & \((0,184)\) & \((1,-1)\) \\
B2 & \(b_2\!\to f\) & 18 & \((0,198)\) & \((1,-1)\) \\
B3 & \(b_3\!\to f\) & 16 & \((0,197)\) & \((1,-1)\) \\
B4 & \(b_4\!\to f\) & 13 & \((0,158)\) & \((0,0)\) \\
\bottomrule
\end{tabular}
\caption{Candidate-B worlds and seed-2026 counts. Link labels and weights are
given in Table~\ref{tab:short_arct_links}.}
\label{tab:short_arct_worlds_b}
\end{table}

For example, in A4 the hard source formula forces \(a_5\) true. In the
lower-cost forced optimum, the weight-\(100\) exact link is preserved, making
\(c\) true; the hard target then requires \(\neg f\), so the active entailment
clause \([\neg a_5\lor f]_{42}\) is violated. Hence
\(\rho_{A4}=42/142-0=0.295775\).  By contrast, A2 supports \(\neg f\) with no
semantic penalty, so \(\rho_{A2}=0/198-1=-1\).

\paragraph{(d) Aggregation and pairwise decision.}

\textsc{Top-Link} selects the highest-confidence link for \(f\), namely
\(a_2\!\to f\) and \(b_2\!\to f\), in both cases contradiction with
\(p=0.982\). The selected Candidate-A and Candidate-B configurations both
have resistance \(-1\), so \textsc{Top-Link} ties. Exact marginalization
instead gives
\[
\begin{aligned}
\mu_A
 &=\tfrac15(0-1-1-1+0.295775)=-0.540845,\\
\mu_B
 &=\tfrac15(0-1-1-1+0)=-0.600000,
\end{aligned}
\]
and therefore selects B because lower resistance is preferred.  The
seed-2026 Monte Carlo estimate uses the counts in
Tables~\ref{tab:short_arct_worlds_a} and \ref{tab:short_arct_worlds_b}:
\[
\begin{aligned}
\widehat\mu_A
 &=\frac{17(-1)+17(-1)+17(-1)}{100}\\
 &\quad+\frac{27(0.295775)+22(0)}{100}
   =-0.430141,\\
\widehat\mu_B
 &=\frac{29(-1)+18(-1)+16(-1)}{100}\\
 &\quad+\frac{13(0)+24(0)}{100}
   =-0.630000.
\end{aligned}
\]
Thus sampled PWAL, exact PWAL, and all ten evaluation
seeds select the gold Candidate B, while \textsc{Top-Link} ties.
\FloatBarrier
\section{Detailed Stable \(\alpha\)NLI Failure Trace}
\label{sec:audited_failure_trace}

In this fully enumerable \(\alpha\)NLI instance, \textsc{Top-Link} selects the
correct candidate, whereas PWAL selects the wrong candidate for every seed
2026--2035 and under exact marginalization. Candidate A is gold. It is
illustrative only and was not used for method or setting selection.
Tables~\ref{tab:short_anli_failure_input}--\ref{tab:short_anli_representative_worlds}
report the input, atoms, links,
categorical choices, and representative worlds.

\paragraph{(a) Input, fixed translations, and atom meanings.}

\begin{table*}[!t]
\centering
\begin{tabular}{@{}p{0.19\textwidth}p{0.75\textwidth}@{}}
\toprule
Premise & Bill was poor. \\
Candidate A & His business went bankrupt and he had to take a new job. \\
Candidate B & His business is worth millions. \\
Outcome & Bill made less money as a computer scientist than he did before. \\
\bottomrule
\end{tabular}
\caption{Stable \(\alpha\)NLI failure trace. Candidate A is gold.}
\label{tab:short_anli_failure_input}
\end{table*}

The source and target formulae are conjunctions:
\[
\Phi_{S_A}=\bigwedge_{i=1}^{6}a_i,\qquad
\Phi_{S_B}=\bigwedge_{i=1}^{4}b_i,\qquad
\Phi_T=\bigwedge_{j=1}^{7}h_j.
\]

The atom surfaces in Table~\ref{tab:short_anli_atoms} are reproduced verbatim
from the translator output and were not manually edited.

\begin{table*}[!t]
\centering
\begin{tabular}{@{}clclcl@{}}
\toprule
\multicolumn{2}{c}{Candidate A source} &
\multicolumn{2}{c}{Candidate B source} &
\multicolumn{2}{c}{Target} \\
\midrule
\(a_1\) & Bill is poor & \(b_1\) & Bill is poor & \(h_1\) & computer scientist \\
\(a_2\) & business is bankrupted & \(b_2\) & multiple 1000000 dollars & \(h_2\) & Bill make money \\
\(a_3\) & job is new & \(b_3\) & he's business & \(h_3\) & Bill make money \\
\(a_4\) & he's business & \(b_4\) & business worth multiple & \(h_4\) & Bill make toward scientist \\
\(a_5\) & he take job & & & \(h_5\) & Bill make during before \\
\(a_6\) & he obligate take & & & \(h_6\) & money have quant by less \\
& & & & \(h_7\) & money have quant money \\
\bottomrule
\end{tabular}
\caption{Complete atom inventory. \(h_2\) and \(h_3\) are distinct compiled
role occurrences even though their verbalizations coincide.}
\label{tab:short_anli_atoms}
\end{table*}

\FloatBarrier

Unlike the ARCT trace, this instance has no exact surface links: all semantic
links below are uncertain atom-level NLI links.

\paragraph{(b) Complete NLI link inventory.}

As before, \(w(p)=\max\{1,\lfloor100p\rfloor\}\). A retained entailment
\(s\!\to h\) contributes \([\neg s\lor h]_{w(p)}\), and a retained
contradiction contributes \([\neg s\lor\neg h]_{w(p)}\).  Because \(h_2\)
and \(h_3\) have identical candidate sets, a row marked \(h_2,h_3\)
represents two separate links, one to each target occurrence.

\begin{table*}[t]
\centering
\begin{tabular}{@{}ccclcrr@{}}
\toprule
Cand. & Target & Source & Source surface & Label & \(p\) & \(w(p)\) \\
\midrule
A & \(h_2,h_3\) & \(a_1\) & Bill is poor & \texttt{Con} & 0.989 & 98 \\
A & \(h_2,h_3\) & \(a_2\) & business is bankrupted & \texttt{Con} & 0.973 & 97 \\
A & \(h_2,h_3\) & \(a_4\) & he's business & \texttt{Ent} & 0.670 & 67 \\
A & \(h_2,h_3\) & \(a_6\) & he obligate take & \texttt{Ent} & 0.587 & 58 \\
A & \(h_5\) & \(a_3\) & job is new & \texttt{Con} & 0.983 & 98 \\
A & \(h_6\) & \(a_1\) & Bill is poor & \texttt{Ent} & 0.579 & 57 \\
A & \(h_6\) & \(a_2\) & business is bankrupted & \texttt{Ent} & 0.423 & 42 \\
A & \(h_7\) & \(a_1\) & Bill is poor & \texttt{Con} & 0.918 & 91 \\
A & \(h_7\) & \(a_2\) & business is bankrupted & \texttt{Con} & 0.886 & 88 \\
\midrule
B & \(h_2,h_3\) & \(b_1\) & Bill is poor & \texttt{Con} & 0.989 & 98 \\
B & \(h_2,h_3\) & \(b_2\) & multiple 1000000 dollars & \texttt{Ent} & 0.685 & 68 \\
B & \(h_2,h_3\) & \(b_3\) & he's business & \texttt{Ent} & 0.670 & 67 \\
B & \(h_2,h_3\) & \(b_4\) & business worth multiple & \texttt{Ent} & 0.940 & 94 \\
B & \(h_6\) & \(b_1\) & Bill is poor & \texttt{Ent} & 0.579 & 57 \\
B & \(h_6\) & \(b_2\) & multiple 1000000 dollars & \texttt{Con} & 0.551 & 55 \\
B & \(h_6\) & \(b_4\) & business worth multiple & \texttt{Con} & 0.761 & 76 \\
B & \(h_7\) & \(b_1\) & Bill is poor & \texttt{Con} & 0.918 & 91 \\
B & \(h_7\) & \(b_2\) & multiple 1000000 dollars & \texttt{Ent} & 0.901 & 90 \\
B & \(h_7\) & \(b_4\) & business worth multiple & \texttt{Ent} & 0.774 & 77 \\
\bottomrule
\end{tabular}
\caption{All 27 retained NLI links, grouped only when two target occurrences
have identical source, label, and confidence.}
\label{tab:short_anli_links}
\end{table*}

The nine grouped A rows contain 13 links because the first four rows each
occur for both \(h_2\) and \(h_3\).  The ten grouped B rows similarly contain
14 links.  Neutral source--target pairs are absent from the retained
inventory and never enter a world.

\paragraph{(c) Categorical world construction.}

For each target atom, PWAL samples uniformly from its retained non-neutral
links plus an explicit no-link outcome.  Target atoms are sampled
independently.  The complete per-target distributions are therefore:

\begin{table*}[t]
\centering
\begin{tabular}{@{}cp{0.40\textwidth}p{0.40\textwidth}@{}}
\toprule
Target & Candidate A choices & Candidate B choices \\
\midrule
\(h_1\) & no-link (probability 1) & no-link (probability 1) \\
\(h_2\) & no-link, \(a_1\) \texttt{Con}, \(a_2\) \texttt{Con}, \(a_4\) \texttt{Ent}, \(a_6\) \texttt{Ent} (each \(1/5\)) & no-link, \(b_1\) \texttt{Con}, \(b_2\) \texttt{Ent}, \(b_3\) \texttt{Ent}, \(b_4\) \texttt{Ent} (each \(1/5\)) \\
\(h_3\) & same five A choices as \(h_2\) (each \(1/5\)) & same five B choices as \(h_2\) (each \(1/5\)) \\
\(h_4\) & no-link (probability 1) & no-link (probability 1) \\
\(h_5\) & no-link, \(a_3\) \texttt{Con} (each \(1/2\)) & no-link (probability 1) \\
\(h_6\) & no-link, \(a_1\) \texttt{Ent}, \(a_2\) \texttt{Ent} (each \(1/3\)) & no-link, \(b_1\) \texttt{Ent}, \(b_2\) \texttt{Con}, \(b_4\) \texttt{Con} (each \(1/4\)) \\
\(h_7\) & no-link, \(a_1\) \texttt{Con}, \(a_2\) \texttt{Con} (each \(1/3\)) & no-link, \(b_1\) \texttt{Con}, \(b_2\) \texttt{Ent}, \(b_4\) \texttt{Ent} (each \(1/4\)) \\
\bottomrule
\end{tabular}
\caption{Per-target categorical choices. Every listed choice within a cell
has the displayed common probability.}
\label{tab:short_anli_distributions}
\end{table*}

\FloatBarrier

Thus the candidate-specific world counts are
\[
\begin{aligned}
N_A&=1\cdot5\cdot5\cdot1\cdot2\cdot3\cdot3=450,\\
N_B&=1\cdot5\cdot5\cdot1\cdot1\cdot4\cdot4=400.
\end{aligned}
\]
All A worlds have probability \(1/450\), and all B worlds have
probability \(1/400\). The displayed sampled run draws 100 worlds per candidate
with seed 2026 rather than enumerating these world spaces.

\paragraph{(d) Deterministic \textsc{Top-Link} calculation.}

\textsc{Top-Link} activates the highest-confidence non-neutral link for every
target that has one.  Its active sets are
\[
\begin{aligned}
L_A^{\mathrm{top}}={}&
\{a_1\!\to h_2\;(\texttt{Con}),
  a_1\!\to h_3\;(\texttt{Con}),\\
& a_3\!\to h_5\;(\texttt{Con}),
  a_1\!\to h_6\;(\texttt{Ent}),\\
& a_1\!\to h_7\;(\texttt{Con})\},\\
L_B^{\mathrm{top}}={}&
\{b_1\!\to h_2\;(\texttt{Con}),
  b_1\!\to h_3\;(\texttt{Con}),\\
& b_4\!\to h_6\;(\texttt{Con}),
  b_1\!\to h_7\;(\texttt{Con})\}.
\end{aligned}
\]
For A, the active capacity is
\(C_{\max}=98+98+98+57+91=442\).  Forcing the positive target conjunction
violates the four contradiction clauses (cost
\(98+98+98+91=385\)), while the entailment witnesses one of seven target
clauses.  Hence
\[
\rho_A^{\mathrm{top}}=385/442-1/7=0.728184.
\]
For B, all four selected links are contradictions, so
\(\Delta T=C_{\max}=98+98+76+91=363\) and no positive target clause is
witnessed:
\[
\rho_B^{\mathrm{top}}=363/363-0=1.
\]
Lower resistance is preferred; \textsc{Top-Link} therefore selects the gold
Candidate A.

\paragraph{(e) PWAL MaxSAT worlds and representative arithmetic.}

Every sampled or enumerated world uses the same base/forced MaxSAT definition
as Section~\ref{sec:supp_constructed_example}:
\[
\rho=\frac{\Delta T}{C_{\max}}-R_{\mathrm{sat}}.
\]
Candidate A and Candidate B have 450 and 400 worlds, respectively. The two
candidate-specific spaces are marginalized separately rather than crossed
into \(450\times400\) joint worlds.
Table~\ref{tab:short_anli_distributions} defines both complete world spaces;
Table~\ref{tab:short_anli_representative_worlds} gives four seed-2026 draws
that illustrate the most tense A world, the worlds closest to the two sampled
means, and the B world with the largest target-clause witness ratio. Part~(f)
aggregates all 100 worlds sampled separately for each candidate.

\begin{table*}[t]
\centering
\begin{tabular}{@{}ccp{0.53\textwidth}rrrr@{}}
\toprule
Cand. & Draw & Active links & \(\Delta T\) & \(C_{\max}\) &
\(R_{\mathrm{sat}}\) & \(\rho\) \\
\midrule
A & 86 & \(a_2\!\to h_3\) \texttt{Con} & 97 & 97 & 0 & 1.000000 \\
A & 76 & \(a_1\!\to h_2\) \texttt{Con}; \(a_6\!\to h_3\) \texttt{Ent};
\(a_3\!\to h_5\) \texttt{Con}; \(a_2\!\to h_6\) \texttt{Ent};
\(a_1\!\to h_7\) \texttt{Con} & 287 & 387 & \(2/7\) & 0.455888 \\
B & 12 & \(b_4\!\to h_2\) \texttt{Ent}; \(b_3\!\to h_3\) \texttt{Ent};
\(b_1\!\to h_7\) \texttt{Con} & 91 & 252 & \(2/7\) & 0.075397 \\
B & 20 & \(b_4\!\to h_2\) \texttt{Ent}; \(b_4\!\to h_3\) \texttt{Ent};
\(b_1\!\to h_6\) \texttt{Ent}; \(b_4\!\to h_7\) \texttt{Ent} & 0 & 322 & \(4/7\) &
-0.571429 \\
\bottomrule
\end{tabular}
\caption{Representative seed-2026 candidate-specific worlds for the stable
\(\alpha\)NLI failure trace. \texttt{Con} and \texttt{Ent} refer to
Table~\ref{tab:short_anli_links}. Draw is the index within the fixed
seed-2026 candidate-specific sample.}
\label{tab:short_anli_representative_worlds}
\end{table*}

\FloatBarrier

For example, draw A76 has three contradiction clauses whose weights sum to
\(98+98+91=287\); those clauses are violated when the positive target is
forced.  Its two entailment links witness \(h_3\) and \(h_6\), giving
\[
\rho_{A76}=287/387-2/7=0.455888.
\]
In B20 all four active links entail distinct positive target occurrences, so
there is no semantic penalty and four of seven clauses are witnessed:
\(\rho_{B20}=0/322-4/7=-0.571429\). These rows illustrate why many B worlds
receive lower resistance than A worlds.

\paragraph{(f) Monte Carlo aggregation, exact check, and decision.}

For the \(K=100\) seed-2026 worlds sampled separately for each candidate, let
\[
\overline T=\frac{1}{K}\sum_{k=1}^{K}
\frac{\Delta T_k}{C_{\max,k}},\qquad
\overline R=\frac{1}{K}\sum_{k=1}^{K}R_{\mathrm{sat},k}.
\]
Then \(\widehat\mu=\overline T-\overline R\), and the seed-2026 sample
averages are
\[
\begin{array}{c@{\qquad}ccc}
 & \overline T & \overline R &
 \widehat\mu=\overline T-\overline R\\
A & 0.667264 & 0.204286 & 0.462978\\
B & 0.373797 & 0.284286 & 0.089512
\end{array}
\]
PWAL therefore selects B, the distractor. Exact marginalization over all
450 Candidate-A worlds and 400 Candidate-B worlds gives
\[
\begin{aligned}
\mu_A&=0.449152,\\
\mu_B&=0.098678,\\
\mu_B-\mu_A&=-0.350474.
\end{aligned}
\]
Exact marginalization therefore confirms the same wrong decision. The
absolute exact margin is \(0.350474\); all
seeds 2026--2035 choose B, and the smallest sampled absolute margin is
\(0.2917\). Thus the observed failure is separated from the tie threshold and
persists across all ten evaluation seeds as well as under exact
marginalization.

The arithmetic also localizes the failure mechanism at the score level.  In
the seed-2026 sample the distractor has both lower normalized semantic tension
(0.3738 versus 0.6673) and a larger target-clause witness ratio
(0.2843 versus 0.2043).
This trace does not, by itself, assign the upstream cause to AMR parsing,
formula compilation, atom-level NLI, or the resistance function.

\end{document}